\documentclass{article}

\usepackage{microtype}
\usepackage{graphicx}
\usepackage{booktabs} 

\usepackage{hyperref}

\usepackage[accepted]{icml2025}

\usepackage[caption=false,font=footnotesize]{subfig}

\usepackage[algo2e,linesnumbered, ruled]{algorithm2e} 

\usepackage{amsmath}
\usepackage{amssymb}
\usepackage{mathtools}
\usepackage{amsthm}

\usepackage[capitalize,noabbrev]{cleveref}

\usepackage{url}            
\usepackage{amsfonts}       
\usepackage{nicefrac}       
\usepackage{xcolor}         
\usepackage{lipsum}

\usepackage{graphicx} 
\usepackage{amsmath}
\usepackage{amssymb}
\usepackage{multirow}
\usepackage{tabularx}
\usepackage{threeparttable}

\newcommand{\mc}[1]{\mathcal{#1}}

\let\oldequation\equation
\let\endoldequation\endequation
\renewenvironment{equation}{%
  \begin{small} 
  \oldequation
}{%
  \endoldequation
  \end{small}
}

\theoremstyle{plain}

\theoremstyle{definition}

\theoremstyle{remark}

\usepackage[textsize=tiny]{todonotes}

\icmltitlerunning{Generalized Phase Pressure Control Enhanced RL for Traffic Signal Control}

\begin{document}

\twocolumn[
\icmltitle{Generalized Phase Pressure Control Enhanced \\ Reinforcement Learning for Traffic Signal Control}




\begin{icmlauthorlist}
\icmlauthor{Xiao-Cheng Liao}{vuw}
\icmlauthor{Yi Mei}{vuw}
\icmlauthor{Mengjie Zhang}{vuw}
\icmlauthor{Xiang-Ling Chen}{vuw}
\end{icmlauthorlist}

\icmlaffiliation{vuw}{Centre for Data Science and Artificial Intelligence \& School of Engineering and Computer Science, Victoria University of Wellington, Wellington, New Zealand}

\icmlcorrespondingauthor{Xiao-Cheng Liao}{xiaocheng@ecs.vuw.ac.nz}

\icmlkeywords{traffic signal control, reinforcement learning, traffic state representation, signalized intersections, transportation}

\vskip 0.3in
]



\printAffiliationsAndNotice{}  

\begin{abstract}
Appropriate traffic state representation is crucial for learning traffic signal control policies.
However, most of the current traffic state representations are heuristically designed, with insufficient theoretical support.
In this paper, we (1) develop a flexible, efficient, and theoretically grounded method, namely generalized phase pressure (G2P) control, which takes only simple lane features into consideration to decide which phase to be actuated; 2) extend the pressure control theory to a general form for multi-homogeneous-lane road networks based on queueing theory; (3) design a new traffic state representation based on the generalized phase state features from G2P control; and 4) develop a reinforcement learning (RL)-based algorithm template named G2P-XLight, and two RL algorithms, G2P-MPLight and G2P-CoLight, by combining the generalized phase state representation with MPLight and CoLight, two well-performed RL methods for learning traffic signal control policies. Extensive experiments conducted on multiple real-world datasets demonstrate that G2P control outperforms the state-of-the-art (SOTA) heuristic method in the transportation field and other recent human-designed heuristic methods; and that the newly proposed G2P-XLight significantly outperforms SOTA learning-based approaches.
Our code is available online\footnote{https://github.com/Rabbytr/g2p-tsc}.
\end{abstract}

\section{Introduction}

Traffic signals, positioned at signalized intersections, play a crucial role in regulating traffic flow across different directions, thereby substantially enhancing both transportation efficiency and road safety.
However, effectively overseeing traffic signals in intricate and dynamic urban road networks remains a formidable undertaking.

Traditional methods to control traffic signal primarily concentrate on developing efficient heuristic algorithms \cite{wang2024llm}, dynamically adjusting traffic light configurations based on traffic conditions at intersections \cite{varaiya2013max}.
However, these methods heavily rely on manual design, demanding substantial human effort and domain knowledge \cite{lai2023large}, which limits the ability to learn from traffic data \cite{gu2024pi}.

With the increasing accessibility of transportation infrastructure such as surveillance cameras, road sensors enable the collection and analysis of extensive real-time data on crucial vehicle and road conditions \cite{lee2021vanet}.
Thus, there is an increasing inclination towards utilizing Deep Reinforcement Learning (DRL) \cite{mnih2015human,sutton2018reinforcement} to fully leverage those real-time data to address the Traffic Signal Control (TSC) problem.
In DRL-based methods, agent(s) can learn traffic signal control strategies via interactions with the environment, resulting in superior performance to traditional approaches \cite{noaeen2022reinforcement,da2024prompt}.

The design of traffic state, reward and neural network architecture all play essential roles in DRL \cite{hu2020learning}.
Some studies \cite{chu2021traffic,wei2018intellilight,van2016coordinated} employ visual images to depict the complete traffic situation at the intersection, resulting in a high-dimensional state with thousands of features.
However, the inclination towards employing more intricate state representations or additional information 
is not always beneficial.
Recent research \cite{zheng2019diagnosing} demonstrated that merely the current number of vehicles on lanes and the current traffic signal phase are sufficient to fully describe the overall system.
The success of effective running state \cite{zhang2022expression} and efficient pressure \cite{wu2021efficient} has demonstrated that significant improvements can be achieved by designing, combining, and optimizing traffic state representations based on simple traffic features with DRL-based models.
The heuristic features designed in these methods have been shown to further improve the performance of RL-based algorithms for TSC.
Despite these studies, current research on heuristic traffic state representations for TSC still faces limitations that can be summarized as follows:

\begin{enumerate}
    \item The current pressure-based state representations, which is typically calculated based on the relative difference between upstream and downstream queues, overlooks the absolute traffic conditions at the intersection. The state-of-the-art (SOTA) algorithms for TSC still have room for improvement by developing more effective traffic state representations.

    \item The existing heuristic-based traffic state representations lack sufficient theoretical justification, which potentially leads to unstable control.

    \item The development of pressure theory is insufficient in multi-lane and multi-homogeneous-lane settings, which are more common in real-world traffic scenarios, and thus require further advancements.
\end{enumerate}

In this paper, we aim to address the above limitations of the existing heuristically designed state representations.
Our core contributions are as follows:
\begin{enumerate}
  \item We present a new method called generalized phase pressure (G2P) control based on the newly designed intersection traffic condition representation, accounting for traffic conditions on both incoming and outgoing roads, as well as the absolute traffic state. We demonstrate G2P control outperforms the SOTA heuristic-based methods.

  \item We extend the pressure control theory to a general form for multi-homogeneous-lane road networks based on queueing theory, and prove that the proposed G2P control is capable of keeping queues of vehicles bounded.

  \item We develop an RL-based algorithm template called G2P-XLight, which combines the generalized turn movement pressure with other learning-based methods and generate two algorithms: G2P-MPLight and G2P-CoLight. Experiment results show that our algorithms are significantly better than SOTA TSC methods.
\end{enumerate}

\section{Background}

\subsection{Preliminaries}
In this section, we summarize relevant definitions for TSC in this paper.
For interested readers, more detailed problem description for TSC can be found in previous research \cite{zheng2019learning}.

\textbf{Traffic network}
A traffic network, also called a road network, comprises a series of roads $\{\mc{R}_0, \mc{R}_1, \mc{R}_2, \cdots \}$ and intersections $\{\mc{I}_0, \mc{I}_1, \mc{I}_2, \cdots \}$ interconnected to establish a comprehensive transportation system facilitating vehicular and pedestrian movements within a geographical area. 

\textbf{Road} 
Each road $\mc{R}_r = \{ \mc{L}_r^{\Lsh}, \mc{L}_r^{\uparrow}, \mc{L}_r^{\Rsh} \}$ typically consists of three types of lanes that dictate how vehicles navigate through intersections, including left-turn lane set $\mc{L}_r^{\Lsh}$, go-straight lane set $\mc{L}_r^{\uparrow}$, and right-turn lane set $\mc{L}_r^{\Rsh}$. 
Lanes in the same set are called homogeneous lanes because all vehicles on them have the same turning direction at the intersection.

\textbf{Intersection}
An intersection $\mc{I}$ is a junction where two or more roads converge or cross. 
At an intersection, roads can be categorized into two types: incoming roads and outgoing roads. 
Incoming roads are where vehicles enter an intersection, and outgoing roads are where they leave the intersection.

\textbf{Traffic turn movement}
A traffic turn movement (TM) $\mc{T}_j = (\mc{R}^{\text{in}}_j, \mc{R}^{\text{out}}_j)$ at an intersection refers to the traffic traveling from one incoming road $\mc{R}^{\text{in}}_j$ to an outgoing road $\mc{R}^{\text{out}}_j$. 
We assume that traffic must travel on the right side and that U-turns are not allowed \cite{wu2021efficient,zhang2022expression}.
According to typical traffic regulations, only vehicles on a specific type of homogeneous lanes (denoted as $\mc{L}^{\Rrightarrow}_j$ in this work) on the incoming road $\mc{R}^{\text{in}}_j$ can cross the intersection from $\mc{R}^{\text{in}}_j$ and $\mc{R}^{\text{out}}_j$.
For example, in a left-turn TM $\mc{T}_j^{\Lsh}$, only vehicles on lanes $\mc{L}_j^{\Rsh} \in \mc{R}^{\text{in}}_j$ are involved. In this case, $\mc{L}^{\Rrightarrow}_j$ is equivalent to $\mc{L}_j^{\Rsh}$.
Similarly, in a go-straight TM $\mc{T}_j^{\rightarrow}$, $\mc{L}^{\Rrightarrow}_j$ is equivalent to $\mc{L}_j^{\uparrow}$.

\textbf{Traffic signal phase}
Some signals cannot turn green simultaneously due to potential conflicts in vehicle movements of different TMs.
At an intersection, there are eight possible pairs of non-conflicting signals \cite{zheng2019learning,buckholz2023introduction}, known as traffic signal phases $\mc{S} = \{s_1,s_2,...,s_8\}$. 
A traffic signal phase $s_i = \{\mc{T}_1, \mc{T}_2\}$ refers to a collection of two permitted TMs.
During a signal phase, all the signals for the TMs associated with that phase turn green, allowing the corresponding vehicles to cross the intersection, while all other TMs are given a red light, ensuring safe and orderly traffic flow.

\textbf{Multi-intersection traffic signal control}
Effective traffic signal control is crucial for preventing accidents, minimizing congestion, and maintaining smooth traffic flow.
At any given intersection, when a traffic signal transition is needed, the learned policy should determine the optimal phase to actuate based on the current traffic state.
Given the time period $T$ of analysis, the objective is to minimize the average travel time of all vehicles spent between entering and leaving area in the traffic network.

\subsection{Related Work}

Recently, the remarkable success of RL-based TSC methods \cite{wei2021recent} can be credited to two key factors:
1) the utilization of advanced deep neural networks for modeling the TSC problem;
2) the development of well-designed traffic state representations and reward systems \cite{chen2024learning,wei2019presslight}.

FRAP \cite{zheng2019learning} introduces an innovative network structure based on phase competition and relations, which effectively manages unbalanced traffic flow and ensures invariance to symmetry properties.
CoLight \cite{wei2019colight} models the communication among intersections based graph attention networks and achieves intersection level cooperation.
MPLight \cite{chen2020toward} bases the neural network architecture on FRAP and implements city-level traffic signal control in a decentralized manner \cite{el2013multiagent,arel2010reinforcement} with parameter sharing.
PressLight \cite{wei2019presslight} designs the RL reward based on the concept of pressure and demonstrates that the theoretically grounded reward design can lead to exceptional performance.

Most TSC methods rely on feedback signal control based on the real-time traffic conditions \cite{xu2021hierarchically,liu2017cooperative,liao2024learning}.
Some methods \cite{chu2021traffic,wei2018intellilight} utilize the visual images of the intersection directly as observations for the RL agent.
This type of state representation can result in thousands of dimensions, necessitating a vast number of training samples.
Recent studies \cite{zheng2019diagnosing,chen2020toward} demonstrate that merely the number of queuing vehicles on lanes and the current traffic signal phase are sufficient to find a good policy for TSC.
On this empirical basis, efficient pressure \cite{wu2021efficient} is heuristically designed as the average difference of queue length between the upstream and downstream.
Utilizing efficient pressure as the traffic state can enhance the performance of RL-based methods.
Besides, advanced traffic state \cite{zhang2022expression} is designed as a combination of efficient pressure and the count of vehicles currently in motion on upstream lanes.
By incorporating the advanced traffic state into CoLight, Advanced-CoLight achieves SOTA performance in learning-based methods and Advanced-MP achieves the new SOTA in heuristic methods.
The traffic state representation significantly influences the quality of TSC models \cite{zheng2019diagnosing,van2016coordinated}.
However, the existing methods still have limitations, such as lack of reflection of absolute traffic conditions and necessary theoretical justification.

In this work, we aim to address these limitations by designing more effective and theoretically grounded traffic state representation.
Consequetly, the following research questions are to be answered:
\textbf{RQ1}) is it possible to design an entirely new pressure-based algorithm that outperforms Advanced-MP?
\textbf{RQ2}) how to represent the traffic state with consideration of absolute traffic conditions and ensure it is theoretically supported?
and \textbf{RQ3}) how to further improve the performance of RL-based methods without significantly increasing the algorithm complexity?

\section{Method}

In this section, we begin by outlining the generalized phase pressure (G2P) control method.
Following this, the theoretical justification for G2P is presented. 
Finally, we integrate G2P into RL-based models and devise the corresponding G2P-XLight algorithms.

\subsection{Generalized Phase Pressure (G2P) Control}

In order to better describe the proposed G2P, we first design the following new concepts:

\textbf{Queue length}
The traffic dynamics within a signalized road network can be represented as a network of vehicle queues \cite{varaiya2013max}. 
Specifically, the road network state is defined by a vector comprising the queue lengths at the intersections of the road network.
The vehicle queue length is typically calculated based on the smallest unit of vehicle movement, the lane $l$.
The queue length of lane $l$ is denoted as $Q(l)$ and can be calculated by:
\begin{equation}
    Q\left(l\right) = \left| \left\{ \mc{C}_k \mid \mc{C}_k \Subset l \right \} \right|,
\end{equation}%
where $\mc{C}_k \Subset l$ indicates that a vehicle $\mc{C}_k$ queuing on $l$.

In a scenario with multiple homogeneous lanes, i.e., $|\mc{L}| > 1$, since vehicles on homogeneous lanes all turn into the same outgoing road at the intersection, every vehicle queue on lane $l \in \mc{L}$ will eventually transfer to this outgoing road.
Therefore, in this work, we define the actual vehicle queue length for a set of homogeneous lanes to be the sum of the queue lengths of vehicles on all relevant homogeneous lanes:
\begin{equation}
    Q\left( \mc{L} \right) = \sum_{l \in \mc{L}} Q\left(l\right),
\end{equation}%
where $Q(\mc{L})$ denotes the actual queue length of a set of lanes.


\textbf{Turn probability}
All vehicles on homogeneous incoming lanes $\mc{L}^{\Rrightarrow}$ will cross the intersection and enter the same outgoing road. 
However, they will be distributed across different types of lanes (i.e., $\mc{L}^{\Lsh}$, $\mc{L}^{\uparrow}$ and $\mc{L}^{\Rsh}$) on that outgoing road.
To represent this mathematically, we introduce a concept of turn probability $P(\mc{L} \mid \mc{L}^{\Rrightarrow})$. 
For example, $P(\mc{L}^{\Lsh} \mid \mc{L}^{\Rrightarrow})$ is the turn probability of vehicles on $\mc{L}^{\Rrightarrow}$ cross the intersection and enter the left-turn lanes of $\mc{L}^{\Lsh}$.

\textbf{Truncated queue length}
Before introducing truncated queue length, the concept of effective range \cite{zhang2022expression} is presented firstly.
Effective range $L_{\text{max}}$ refers to the maximum distance that vehicles can travel to reach the intersection before the next signal change. It can be calculated by:
\begin{equation}
    L^{\text{max}}_r = \min\left( V^{\text{max}}_r, V^{\text{max}}_c \right) \Delta t,
\end{equation}%
where $V^{\text{max}}_c$ and $V^{\text{max}}_r$ are the maximum speed of vehicles and the maximum speed limit of the road, respectively.
$\Delta t$ is the duration of the actuated signal phase.

\begin{figure}[!t]
\centering
\includegraphics[width=\columnwidth]{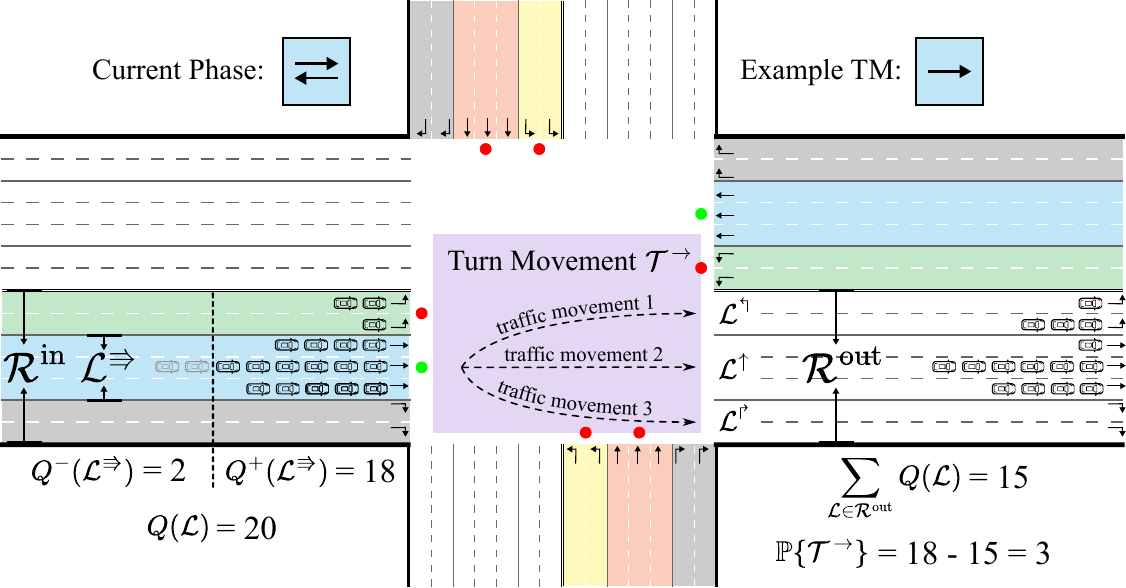}
\caption{A multi-homogeneous-lane intersection, where a single road has multiple lanes with the same turning direction, is shown with illustrations of some symbols and concepts as well as an example of calculating the generalized turn movement pressure $\mathbb{P}\{\mc{T}^{\rightarrow}\}$.}
\label{fig_g2p}
\end{figure}

Based on the effective range, we introduce the new concept of truncated vehicle queue.
The truncated vehicle queue refers to the portion of the queue in the incoming lanes that is within the effective range, its length can be calculated by:
\begin{equation}
    Q^+\left( \mc{L}^{\Rrightarrow} \right) = 
    \sum_{l\in \mc{L}^{\Rrightarrow}} 
    \left| \left\{ \mc{C}_k \mid D\left(\mc{C}_k, \mc{I} \right) \leq L_{\text{max}} 
    \land \mc{C}_k \Subset l
    \right\} \right|,
\end{equation}%
where $D(\mc{C}_k, \mc{I})$ denotes the distance between vehicle $\mc{C}_k$ and the intersection $\mc{I}$,
$L_{\text{max}}$ is the corresponding effective range of lane $l$.
An illustration of the truncated queue length is presented in Figure \ref{fig_g2p}.

\textbf{Traffic movement pressure}
Traffic movement refers to the movement of vehicles from homogeneous lanes $\mc{L}^{\Rrightarrow}$ on the incoming road to a specific homogeneous lane set $\mc{L}$ on the outgoing road, which can be represented as a pair $(\mc{L}^{\Rrightarrow}, \mc{L})$, where $\mc{L} \in \mc{R}^{\text{out}}$.

Intuitively, a traffic movement with higher (lower) pressure should be more (less) urgent to be actuated in order to reduce the average travel time of vehicles in the road network. 
A reasonable traffic movement pressure should consider the three factors: 1) take into account the traffic conditions of the incoming lanes; 2) take into consideration the traffic conditions of the outgoing lanes; and 3) reflect the absolute traffic conditions rather than relative states.

Based on the above three considerations, we base absolute limit on the operator $\min(\cdot)$ and design the traffic movement pressure as follows:
\begin{equation}
    \mathbb{P}\left\{ \left(\mc{L}^{\Rrightarrow}, \mc{L} \right) \right\} =
    \min\left( 
        P\left(\mc{L} \mid \mc{L}^{\Rrightarrow}\right)
        Q\left(\mc{L}^{\Rrightarrow}\right),
        \beta \right)
    - Q\left(\mc{L}\right),
\end{equation}%
where $\beta$ is proposed to be a value determined by the specific configuration of the corresponding involved roads and intersection.
The motivation for incorporating term $\beta$ into the pressure is to ensure that the pressure is able to reflect the absolute traffic conditions to some extent.
When the vehicle queues on both the incoming lanes and the corresponding outgoing lanes are long, a relative value may fail to accurately capture the pressure of that traffic movement.
In this case, selecting an appropriate value for $\beta$ is crucial.
In this work, we propose to design $\beta$ based on the truncated queue from incoming lanes:
We propose to set $\beta$ as the number of vehicles in the truncated queue from incoming lanes that are expected to enter the lanes $\mc{L}$:
\begin{equation}
\begin{aligned}
    \mathbb{P}\left\{ \left(\mc{L}^{\Rrightarrow}, \mc{L} \right) \right\} &=
    \min\bigg(
        P\left(\mc{L} \mid \mc{L}^{\Rrightarrow}\right)
        Q\left(\mc{L}^{\Rrightarrow}\right),\\
        &P^+\left(\mc{L} \mid \mc{L}^{\Rrightarrow}\right)
        Q^+\left(\mc{L}^{\Rrightarrow}\right) 
        \bigg)
    - Q\left(\mc{L}\right),
\end{aligned}
\end{equation}%
where $P^+(\mc{L} \mid \mc{L}^{\Rrightarrow})$ denotes the turn probability of vehicles within the truncated queue of $\mc{L}^{\Rrightarrow}$ cross the intersection and enter the lanes of $\mc{L}$ in the outgoing road.
The rationale for setting $\beta$ this way is that only vehicles in the truncated queue are likely to be directly affected by the current phase.
Since $\forall \mc{L} \in \mc{R}^{\text{out}}$,
\begin{equation}
    P\left(\mc{L} \mid \mc{L}^{\Rrightarrow}\right)
        Q\left(\mc{L}^{\Rrightarrow}\right) \geq
        P^+\left(\mc{L} \mid \mc{L}^{\Rrightarrow}\right)
        Q^+\left(\mc{L}^{\Rrightarrow}\right) 
\end{equation}%
always holds. 
Therefore, the calculation for the traffic movement pressure can be simplified to:
\begin{equation}
    \mathbb{P}\left\{ \left(\mc{L}^{\Rrightarrow}, \mc{L} \right) \right\} =
    P^+\left(\mc{L} \mid \mc{L}^{\Rrightarrow}\right)
    Q^+\left(\mc{L}^{\Rrightarrow}\right) - Q\left(\mc{L}\right).
    \label{eq:traffic_mm_pressure}
\end{equation}%

\subsection{Generalized Turn Movement Pressure}
Based on the traffic movement pressure defined above, which defines lane-level pressure, we can further extend it to generalized road-level pressure, referred to as generalized turn movement pressure:
\begin{equation}
\begin{aligned}
    \mathbb{P}\left\{ \mc{T}_j \right\} &=
    \sum_{ \mc{L} \in \mc{R}_j^{\text{out}} }
    \mathbb{P}\left\{ \left(\mc{L}^{\Rrightarrow}, \mc{L} \right) \right\} \\
    & = \sum_{ \mc{L} \in \mc{R}_j^{\text{out}} }
    \left[
    P^+\left( \mc{L} \mid \mc{L}^{\Rrightarrow}_j \right)
    Q^+\left( \mc{L}^{\Rrightarrow}_j \right) - Q\left(\mc{L}\right)
    \right]\\
    & = Q^+\left( \mc{L}^{\Rrightarrow}_j \right)
    \sum_{ \mc{L} \in \mc{R}_j^{\text{out}} } P^+\left( \mc{L} \mid \mc{L}^{\Rrightarrow}_j \right)
    - \sum_{ \mc{L} \in \mc{R}_j^{\text{out}} } Q\left(\mc{L}\right) \\
    & = Q^+\left( \mc{L}^{\Rrightarrow}_j \right)
    - \sum_{ \mc{L} \in \mc{R}_j^{\text{out}} } Q\left(\mc{L}\right) \\
    & = Q\left(\mc{L}^{\Rrightarrow}_j\right)
    -Q^-\left( \mc{L}^{\Rrightarrow}_j \right)
    - \sum_{ \mc{L} \in \mc{R}_j^{\text{out}} } Q\left(\mc{L}\right),
\end{aligned}
\label{eq:tm_pr_calculation}
\end{equation}%
where $Q^-(\mc{L}^{\Rrightarrow}_j)$ is the residual queue length which is defined as:
\begin{equation}
    Q^-\left( \mc{L}^{\Rrightarrow} \right) = 
    \sum_{l\in \mc{L}^{\Rrightarrow}} 
    \left| \left\{ \mc{C}_k \mid D\left(\mc{C}_k, \mc{I} \right) > L_{\text{max}} 
    \land \mc{C}_k \Subset l
    \right\} \right|.
\end{equation}%
According to \eqref{eq:tm_pr_calculation}, the value of any turn probability is not required for calculating the generalized turn movement pressure, which is of importance for real-world application.
To facilitate reader understanding, an example to calculate $\mathbb{P}\{ \mc{T}_j\}$ is presented in Figure \ref{fig_g2p}.

\subsection{Generalized Phase Pressure (G2P)}
The generalized phase pressure is defined as the sum of generalized TM pressure of two traffic movements from one phase $s_i$, denoted as $\mathbb{P}\{s_i\}$:
\begin{equation}
\mathbb{P}\left\{s_i\right\} = \sum_{\mc{T}_j \in s_i}\mathbb{P}\left\{ \mc{T}_j \right\}.
\label{eq:phase_pr_calculation}
\end{equation}%
The generalized phase pressure accurately reflects the necessity to actuate the corresponding phase by considering the absolute traffic state of both incoming lanes and outgoing roads.
When the $\mathbb{P}\{s_i\}$ is too small, there is likely congestion on the involved outgoing lanes.

\subsection{Algorithm}

Algorithm \ref{alg:g2pcontrol} provides a formal summary of the G2P control. 
At each time when a traffic signal phase transition is needed at any intersection, G2P control first calculates the generalized turn movement pressure $\mathbb{P}\{ \mc{T}_j\}$ for each TM $\mc{T}_j$, and then further computes the generalized phase pressure for each phase $s_i$.
Based on the computed pressures, the method determines the signal phase with the highest generalized phase pressure, denoted as $s^*$, which is selected as the next phase to be actuated. 
Consequently, the corresponding yellow and red lights are actuated to clear the vehicles from the intersection. 
Finally, all signals associated with phase $s^*$ are turned green.
Thus, a complete phase decision process is finished. 
Each time traffic signals of an intersection needs to be changed, the above process will be followed.

\begin{algorithm2e}[!t]
\DontPrintSemicolon
\caption{G2P control}
\label{alg:g2pcontrol}
\For {each time step $t$}{
    \For{each intersection of the road network}{
        \For{each phase $s_i \in \mc{S}$ of the intersection}{
            \For{each TM $\mc{T}_j \in s_i$}{
                Calculate $\mathbb{P}\{ \mc{T}_j\}$ by Eq. \eqref{eq:tm_pr_calculation} \; 
            }
            Calculate $\mathbb{P}\{s_i\}$ by Eq. \eqref{eq:phase_pr_calculation} \;
        }
        $s^* = \mathop{\arg\max}
            \limits_{s_i \in \mc{S}} \mathbb{P}\{s_i\}$ \;
        Set the signal phase of the intersection to $s^*$ \;
    }
    Switch the corresponding red and yellow lights based on changes in signal phase at each intersection \;
}
\end{algorithm2e}

\subsection{G2P Control Stability Analysis}
A key research question is whether the proposed G2P control method can stabilize the queue lengths in traffic.
It is of great significance in real-world applications to guarantee that the algorithm will produce a stable closed-loop system \cite{kouvelas2014maximum}.
Herein, we will prove that the proposed G2P control can stabilize the queue lengths in this subsection.

Let $d_{\mc{L}}$ represents the exogenous flow of vehicles per period in lanes $\mc{L}$, $f_{\mc{L}}$ represents the flow of vehicles per period in lanes $\mc{L}$.
$\mc{R}_{\text{entry}}$ represents the set of entry roads, it means that external vehicles can enter the road network through these roads, 
$\mc{R}_{\text{exit}}$ represents the set of exit roads which means that vehicles can exit the road network through these roads, 
and $\mc{R}_{\text{int}}$ represents the internal roads in the road network.
Based on queuing theory \cite{miller1963settings,anokye2013application}, the following definitions are provided:

\textbf{Law of Routing and flow conservation}:
\begin{equation}
    f_{\mathcal{L}_i} = d_{\mathcal{L}_i}, \mc{L}_i \in \mc{R}_i, \mathcal{R}_i \in \mathcal{R}_{\text{entry}},
    \label{eq_f1}
\end{equation}
\begin{equation}
    f_{\mathcal{L}_j} = \sum_{\mc{L}_j \in \mc{R}_j, \mathcal{R}_j \in \mathcal{R}_{\text{int}}\cup\mathcal{R}_{\text{exit}}} f_{\mathcal{L}_i} P\left(\mc{L}_j \mid \mc{L}_i\right),
    \label{eq_f2}
\end{equation}%
where $P\left(\mc{L}_j \mid \mc{L}_i\right)$ is the proportion of vehicles leaving $\mathcal{L}_i$ and entering $\mathcal{L}_j$.
Flows on entry lanes (connected to external incoming lanes) are determined by exogenous demands: \eqref{eq_f1}. 
Flows on other lanes are governed by routing proportions and flow conservation: \eqref{eq_f2}.

\textbf{Definition 1.} At the end of each period $t$, a signal control matrix $A(t) = \{a(\mathcal{T})(t)\}$ must be selected for use in period $t+1$, where $a(\mathcal{T})$ indicates that the turn movement $\mc{T}$ in this direction is allowed to move safely. 
Therefore, $A(t) = u(Q(t))$ is determined by the feedback control policy, also the admissible phase control sequence $A = \{A(t)\}$ accommodates $f$ (d is a feasible demand) if:
\begin{equation}
\forall \mc{T}, \quad c_{\mathcal{L}^\Rrightarrow}\left(\mathcal{R}^\text{{out}}\right) \Sigma_A\left(\mathcal{T}\right) > f_{\mathcal{L}^\Rrightarrow} 
\end{equation}%
where 
\begin{equation}
\begin{aligned}
& \Sigma_A\left(\mathcal{T}\right) = \lim  \underset{T}{\text{inf}} \frac{1}{T}\sum_{t=1}^T A\left(t\right)\left(\mathcal{T}\right), \\
& \mathcal{L}^{\Rrightarrow} \in \mc{R}^{\text{in}}, \mc{R}^{\text{in}} \in \mathcal{R}_{\text{int}} \cup \mathcal{R}_{\text{entry}}, 
\mc{T} = \left(\mc{R}^\text{{in}}, \mc{R}^\text{{out}}\right).
\end{aligned}
\end{equation}%
This implies that the queue length in homogeneous lanes $\mc{L}^{\Rrightarrow}$ waiting to turn into $\mc{R}^{\text{out}}$ is always bounded.

\textbf{Definition 2.} The lane queue length process $Q(t) = \{Q(\mathcal{L}^{\Rrightarrow})(t)\}$ is considered stable in the mean, and the control policy $u$ is a stabilizing control policy if there exists a constant $K < \infty$:
\begin{equation}
    \forall T, \quad \frac{1}{T} \sum_{t=1}^T \sum_{\mathcal{L}^{\Rrightarrow}} E \left\{ Q\left(\mathcal{L}^{\Rrightarrow}\right)\left(t\right) \right\} < K.
\label{eq_stable}
\end{equation}%
Stability in the mean indicates that the chain is positively recurrent and possesses a unique steady-state probability distribution.
In this equation, $E\{\cdot\}$ denotes expectation.

In our control policy, the control policy $u^*$ is to select the signal phase with the largest generalized phase pressure:
\begin{equation}
    u^*\left(\mc{S}\right) = \mathop{\arg\max}
            \limits_{s_i \in \mc{S}} \mathbb{P}\{s_i\}
    \label{eq_u}
\end{equation}%

Based on the above definitions, we can have the following theorem regarding the G2P stability.

\textbf{Theorem 1.} The G2P control policy $u^*$ is stabilizing whenever the average demand vector d is a feasible demand ($d \in D$, where $D$ is a feasible demand set).

\textit{Proof}: See Appendix \ref{sec:proof}.

\newcommand{\Lrtt}[1]{\rotatebox[origin=c]{#1}{$\Lsh$}}
\newcommand{\Rrtt}[1]{\rotatebox[origin=c]{#1}{$\Rsh$}}
\newcommand{\pressure}[1]{\mathbb{P}\big\{#1\big\}}

\newcommand{\upleft}{\Lsh}
\newcommand{\leftdown}{\Lrtt{90}}
\newcommand{\downright}{\Lrtt{180}}
\newcommand{\rightup}{\Lrtt{-90}}

\newcommand{\upright}{\Rsh}
\newcommand{\rightdown}{\Rrtt{90}}
\newcommand{\downleft}{\Rrtt{180}}
\newcommand{\leftup}{\Rrtt{-90}}

\newcommand{\tm}[1]{\pressure{\mc{T}^{#1}}}

\subsection{G2P-XLight}
Based on the proposed G2P control method, we can create an RL-based algorithm template (as shown in Algorithm \ref{alg:g2pxlight}), referred to as G2P-XLight.
In this work, two SOTA neural network models, MPLight \cite{chen2020toward} and Colight \cite{wei2019colight} are utilized as the basic model for our method, given their respective advantages in convergent stability and strong performance.
Notably, the RL-based design concept is not restricted to only MPLight and CoLight; it can also be integrated into other RL-based models and other learning-based methods.
The following content describes some important components of the proposed XLight.

\begin{algorithm2e}[!t]
\DontPrintSemicolon
\caption{G2P-XLight}
\label{alg:g2pxlight}
\For {each time step $t$}{
    \For{each intersection of the road network}{
        Get the generalized traffic state $O_t^{\text{G2P}}$ \;
        Calculate the next phase $s^*$ by XLight RL model \;
        Set the signal phase of the intersection to $s^*$ \;
    Switch the corresponding red and yellow lights based on changes in signal phase at each intersection \;
    }
}
\end{algorithm2e}

\textbf{State}
According to \eqref{eq:traffic_mm_pressure}, since the turn probability is unknown to the algorithm, we cannot accurately calculate the traffic movement pressure.
Instead, we utilize the traffic turn movement pressure of all TMs in the intersection to construct the traffic state:
\begin{equation}
    O_{\text{TM}} = \left\{ \begin{alignedat}{2}
    &\tm{\downarrow}, \tm{\leftarrow}, \tm{\uparrow}, \tm{\rightarrow},\\
    &\tm{\downright}, \tm{\leftdown}, \tm{\upleft}, \tm{\rightup},\\
    &\tm{\downleft}, \tm{\leftup}, \tm{\upright}, \tm{\rightdown}
    \end{alignedat} \right\},
\end{equation}%
where dimension of $O_{\text{G2P}}$ is determined by the specific road network intersection configuration.
Based on G2P, we design the generalized traffic state by combining the current phase \cite{zheng2019diagnosing}, $O_{\text{TM}}$, and the advanced traffic state \cite{zhang2022expression} together.
In G2P-XLight, at each timestep $t$ each agent observes the newly proposed generalized traffic state $O^{\text{G2P}}_t$ and performs the appropriate actions.

\textbf{Action}
The action space is set as the set of phases for the intersection, that is $\mc{S}$.
At each timestep $t$, each agent selects a traffic signal phase $s^* \in \mc{S}$ as its action $a_t$, and the traffic signal is then set to phase $s^*$.

\textbf{Reward}
For G2P-MPLight algorithm, the reward is defined as $-|\sum_{\forall \mc{L}^{\Rrightarrow} \in \mc{I}} [Q(\mc{L}^{\Rrightarrow}) - \sum_{\forall \mc{L} \in \mc{I}}Q(\mc{L})]|$, which is a theoretically grounded \cite{wei2019presslight} reward design.
The reward for G2P-CoLight is the total length of the intersection $-\sum_{\forall \mc{L}^{\Rrightarrow} \in \mc{I}} Q(\mc{L}^{\Rrightarrow})$.
The goal of G2P-CoLight is to minimize the queue length of the transportation road network.

\section{Experiment Setup}

\makeatletter
\newcommand*{\txtoverline}[1]{$\overline{\hbox{#1}}\m@th$}
\makeatother
\newcommand{\gray}[1]{\textcolor{gray}{#1}}

\begin{table*}[!t]
\tiny 
\renewcommand{\arraystretch}{1} 
\setlength{\tabcolsep}{11pt}
\caption {Performance (the average travel time in seconds) comparison of different algorithms evaluated on 30 independent runs (for learning-based methods) cross real-world datasets (the smaller the better).}
\centering
\begin{tabular}{cllllllll}
\hline\hline
\multicolumn{2}{c|}{\multirow{2}{*}{Method}} 
&\multicolumn{3}{c|}{Jinan} 
&\multicolumn{2}{c|}{Hangzhou} 
&\multicolumn{2}{c}{Manhattan} \\ \cline{3-9}
\multicolumn{2}{c|}{} 
&1 &2 &3 
&1 &2 
&1 &2 \\ \hline

\multicolumn{9}{c}{Traditional methods} \\ \hline

Random       &avg. &722.1711 &644.0502 &674.3132 &690.8250 &641.2051 &1603.7671 &1504.7746  \\ \hline
Fixed-time   &avg. &514.1358 &425.3143 &456.3192 &575.5564 &563.8811 &1582.1029 &1476.5341  \\ \hline
Max-pressure &avg. &374.5641 &328.9629 &332.9403 &365.0634 &446.9059 &1335.7877 &1121.3834  \\ \hline

Efficient-MP &avg. &366.7037 &285.7890 &309.5715 &320.2789 &431.9006 &1369.8349 &1132.3524  \\ \hline
Advance-MP   &avg. &\textbf{352.9977} &282.5613 &303.2879 &317.7901 &428.4886 &1331.5656 &1091.0796  \\ \hline
G2P control  &avg. &353.1016 &\textbf{279.4178} &\textbf{295.6718} &\textbf{311.1354} &\textbf{415.1403} &\textbf{1279.5890} &\textbf{1046.5328}  \\ \hline

\multicolumn{9}{c}{Learning-based methods} \\ \hline

&best 
&327.6437   &304.5425   &303.7150   &327.3208   &416.5344   &1201.2206   &925.3551  
\\MPLight &avg. 
&341.1644(+)&309.3685(+)&314.9360(+)&329.8745(+)&423.1737(+)&1240.7652(+)&961.9344(+)    
\\\cite{chen2020toward}&std. 
&\gray{7.9328}   &\gray{2.3608}   &\gray{7.9433}   &\gray{1.5704}   &\gray{4.2424}   &\gray{32.9315}   &\gray{26.7568}  
\\ \hline
    
&best 
&335.0111   &298.0151   &297.3183   &327.1559   &424.4160   &1201.4507   &928.1192  
\\CoLight &avg. 
&585.2686(+)&330.8309(+)&449.0182(+)&329.3097(+)&478.4843(+)&1354.3869(+)&974.2463(+)    
\\\cite{wei2019colight}&std. 
&\gray{97.3547}   &\gray{88.3939}   &\gray{132.4389}   &\gray{1.3632}   &\gray{46.5520}   &\gray{70.0379}   &\gray{39.5790}  
\\ \hline
    
&best 
&303.4494   &278.5507   &276.2241   &316.2668   &390.8584   &1218.0373   &955.2109  
\\Efficient-MPLight &avg. 
&314.5611(+)&279.6051(+)&277.8913(+)&316.9375(+)&395.2163(+)&1269.4947(+)&1026.0611(+)
\\\cite{wu2021efficient}&std. 
&\gray{5.9581}   &\gray{0.4852}   &\gray{1.1198}   &\gray{0.3454}   &\gray{3.0347}   &\gray{34.9415}   &\gray{45.5581}  
\\ \hline
    
&best 
&334.6210   &280.3205   &285.2552   &315.7187   &413.2984   &1207.1723   &945.4937  
\\Efficient-CoLight &avg. 
&498.5503(+)&281.5098(+)&303.4334(+)&316.2652(+)&418.7666(+)&1280.6021(+)&980.9108(+)
\\\cite{wu2021efficient}&std. 
&\gray{126.8593}   &\gray{0.7810}   &\gray{43.2528}   &\gray{0.2987}   &\gray{4.1263}   &\gray{40.4760}   &\gray{21.8228}  
\\ \hline
    
&best 
&282.4828   &268.1462   &265.4618   &305.8079   &377.3693   &1216.9679   &933.5639  
\\Advanced-MPLight &avg. 
&284.4935(+)&268.6490(+)&266.0353(+)&306.4337(+)&380.8768($\approx$)&1264.5822(+)&1030.1002(+)
\\\cite{zhang2022expression}&std. 
&\gray{1.2856}   &\gray{0.2267}   &\gray{0.4116}   &\bf{\gray{0.1522}}   &\bf{\gray{1.7325}}   &\gray{24.4470}   &\gray{46.4305}  
\\ \hline
    
&best 
&290.0361   &269.9070   &266.6070   &306.1686   &391.7235   &1172.4191   &911.8467  
\\Advanced-CoLight &avg. 
&448.6822(+)&271.2544(+)&291.7819(+)&306.7133(+)&398.2087(+)&1226.5437(+)&942.8266(+) 
\\\cite{zhang2022expression}&std. 
&\gray{114.3992}   &\gray{0.7593}   &\gray{58.8422}   &\gray{0.2148}   &\gray{5.7507}   &\gray{43.3741}   &\bf{\gray{15.4605}}  
\\ \hline

&best 
&\bf{280.9738}   &\bf{267.9407}   &\bf{264.8071}   &\bf{305.5619}   &\bf{375.7645}   &1211.8514   &931.5773  
\\G2P-MPLight &avg. 
&\bf{282.6323}   &\bf{268.4085}   &\bf{265.3298}   &\bf{306.2399}   &\bf{379.8963}   &1247.8534(+)&1010.6038(+)    
\\&std. 
&\bf{\gray{0.8537}}   &\bf{\gray{0.2079}}   &\bf{\gray{0.3866}}   &\gray{0.2628}   &\gray{2.2614}   &\bf{\gray{23.1139}}   &\gray{23.4603}  
\\ \hline
    
&best 
&288.0631   &269.9911   &266.4933   &306.0583   &388.2922   &\bf{1167.2658}   &\bf{897.1371}
\\G2P-CoLight &avg. 
&489.1190(+)&278.7193(+)&337.0601(+)&306.7082(+)&397.0420(+)&\bf{1199.4369}   &\bf{919.0087}    
\\&std. 
&\gray{89.8805}   &\gray{21.5407}   &\gray{82.6323}   &\gray{0.2358}   &\gray{13.2278}   &\gray{27.8526}   &\gray{19.4524}  
\\ \hline
    
\hline
\end{tabular}
\begin{tablenotes}
    \item The best values (including the best value, average, and standard deviation) for each scenario are highlighted.
     \item The symbols "+", "$\approx$", and "-" denote whether the corresponding result is significantly worse than, statistically comparable to, or better than the corresponding highlighted learning-based method.
\end{tablenotes}
\label{table:performance_first3}
\end{table*}

\textbf{Experiment settings}
Our experiments are conducted on CityFlow$\footnote{https://cityflow-project.github.io}$, a traffic simulator which supports large-scale traffic signal control. 
In accordance with tradition \cite{chen2020toward}, each green signal is followed by a three-second yellow light and a two-second all-red interval.
The phase number is set to eight that is the full phase configuration for a complex intersection and the minimum action duration to 10-second \cite{wei2019colight}.

\textbf{Datasets}
In this paper, we take three commonly used real-world traffic datasets\footnote{https://traffic-signal-control.github.io/}: 1) Jinan (comprises three distinct traffic flow datasets); 2) Hangzhou (comprises two distinct traffic flow datasets) and 3) large-scale scenarios in Manhattan (comprises two distinct traffic flow datasets).

\textbf{Evaluation metrics}
Following established research \cite{wei2019colight,zhang2022expression}, we assess the performance of various traffic signal control models by measuring average travel time of vehicles. 

\textbf{Compared methods}
We select five rule-based baselines Random, Fixed-time \cite{koonce2008traffic}, Max-pressure \cite{varaiya2013max}, Efficient-MP \cite{wu2021efficient} and Advanced-MP \cite{zhang2022expression} as baselines for comparison.
Besides, six RL-based methods MPLight \cite{chen2020toward}, CoLight \cite{wei2019colight}, 
Efficient-MPLight \cite{wu2021efficient}, Efficient-CoLight \cite{wu2021efficient}, 
Advanced-MPLight \cite{zhang2022expression} and Advanced-CoLight \cite{zhang2022expression} are utilized as learning-based baselines for our methods.

Notably, Advanced-MPLight and Advanced-CoLight are state-of-the-art methods \cite{gu2024pi} and
all the hyper-parameters of baseline methods in this paper follow the recommendations from their respective the original papers.
The hyper-parameters of G2P-MPLight and G2P-CoLight also follow the recommendations from the original papers of MPLight and CoLight, respectively.
Since random seeds can have a certain impact on RL-based methods, the Wilcoxon rank-sum test \cite{mann1947test} at a significance level of 0.05 is conducted with Bonferroni correction \cite{liao2024learning1} for 30 independent runs.

\section{Results}

\subsection{Overall Performance}
\begin{figure}[!t]
\centering
\includegraphics[width=0.9\columnwidth]{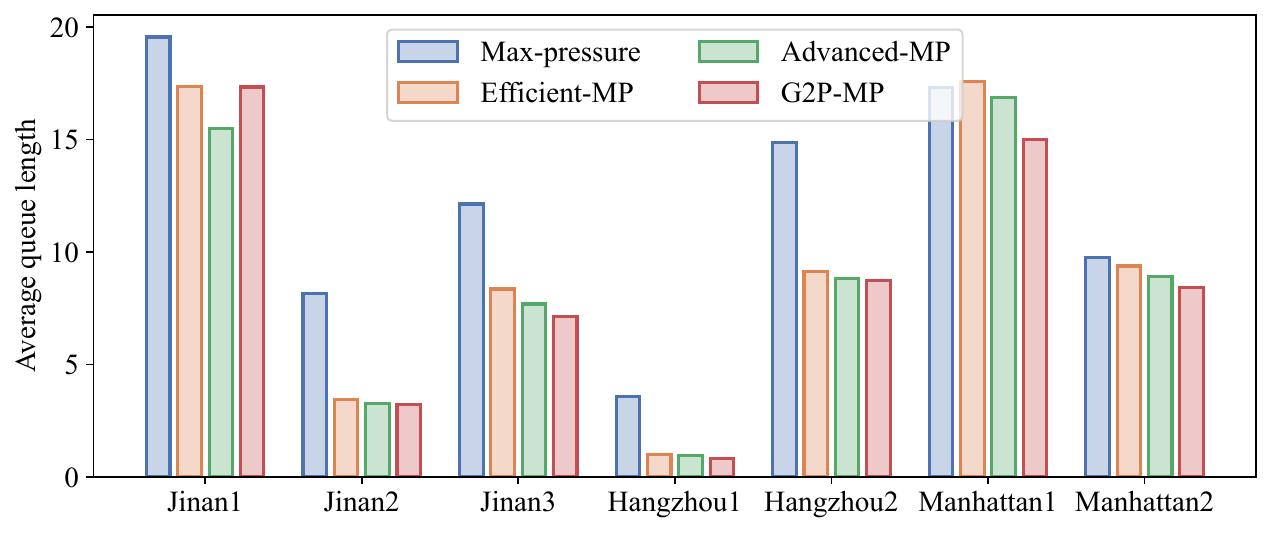}
\caption{The average queue length (the smaller the better) of different algorithms.}
\label{fig:queue_length}
\end{figure}

We demonstrate the performance of heuristic methods as well as RL-based methods on real-world datasets with respect to the average travel time of vehicles in Table \ref{table:performance_first3}.
As can be seen in Table \ref{table:performance_first3},
the proposed G2P control outperforms all the other heuristic methods in almost all scenarios, except for one small-scale scenario in Jinan where G2P control is slightly worse than Advanced-MP.
Although Advanced-MP considers more traffic features, such as vehicle speed, its performance shows a significant gap compared to G2P control in many scenarios.
The superior performance of G2P control compared to Efficient-MP indicates that the proposed generalized phase pressure, which accounts for absolute traffic conditions, is more effective in reducing the average travel time of vehicles than efficient pressure, which only considers relative traffic conditions.
In addition, Figure \ref{fig:queue_length} demonstrates that G2P control can effectively reduce the average queue length of vehicles than other heuristic-based methods in most scenarios.

Promisingly, the performance of G2P control in many scenarios is close to or even surpasses that of some learning-based methods in small-scale scenarios (i.e., Jinan and Hangzhou).
This demonstrates the potential value of G2P.
Compared with Advanced-MP, its implementation requires only a simple and commonly used lane feature, whereas Advanced-MP also requires additional information, such as vehicle speed.
Compared with learning-based methods, G2P control does not need deep models and intensive training, which is easier to be applied in real-world scenarios than learning-based methods.

It can be seen in Table \ref{table:performance_first3} that CoLight-based performs exceedingly well in large-scale scenarios (i.e., scenarios of Manhattan) but it exhibits convergence instability in smaller-scale scenarios, whereas MPLight-based methods demonstrate good convergence stability across all scenarios.
G2P-MPLight significantly outperforms all the baselines in scenarios of Jinan and Hangzhou and G2P-CoLight significantly outperforms all the compared algorithms in scenarios of Manhattan.
This suggests that in large-scale scenarios, communication among intersections (introduced in CoLight) is more important and can significantly improve the algorithm's performance.
However, in small-scale scenarios, the relative complex model of CoLight potentially lead to some convergence instability.
Additionally, G2P-MPLight is always significantly better than Advanced-MPLight.
This demonstrates the effectiveness of the proposed generalized traffic state.

\subsection{Convergence}
The convergence curves of all RL-based comparison algorithms in scenarios Manhattan1 and Manhattan2 are presented in Figure \ref{fig:convergence}.
All the other convergence curves are shown in Appendix \ref{sec:appdx_convergence}, which show the same patterns as Figure \ref{fig:convergence}.
The decentralized manner and relatively simple model in MPLight-based methods exhibit good data efficiency, but without a communication mechanism, it leads to potential performance shortcomings.
CoLight-based methods converge more slowly, but they ultimately achieve much better performance.
This indicates that the communication mechanism introduced by CoLight has a significant impact on the algorithm's performance in large-scale scenarios.
Additionally, by scanning horizontally we can see that the ranking of data efficiency is G2P-CoLight $>$ Efficient-CoLight $>$ Advanced-CoLight.
This indicates that the additional information introduced by advanced traffic state reduces the data efficiency, whereas the generalized traffic state does not have this issue and even achieves higher data efficiency while maintaining algorithm performance.


\begin{figure}[!t]
\centering
\includegraphics[width=0.9\columnwidth]{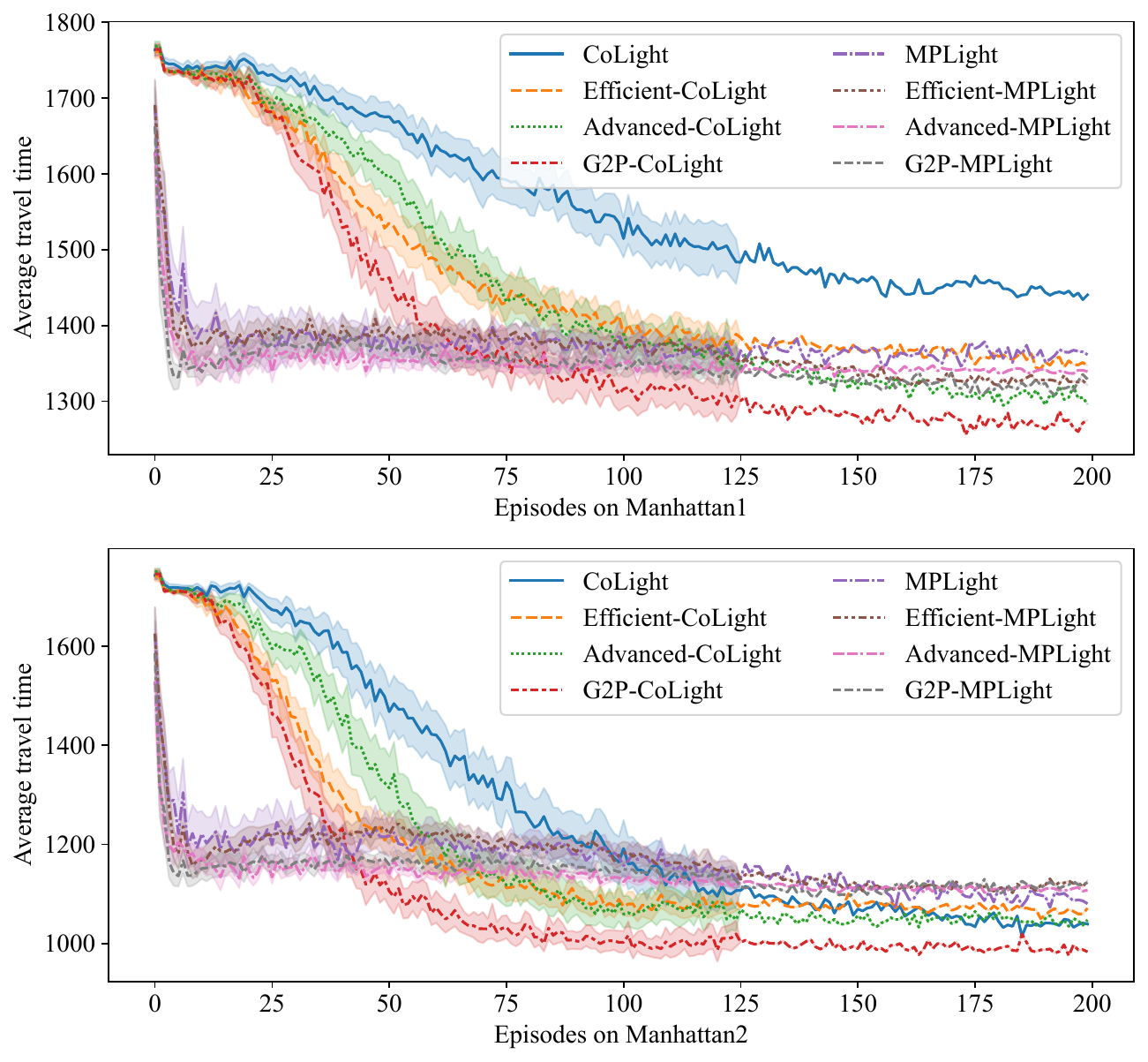}
\caption{For each algorithm we used 30 seeds and obtained the mean-curve with 95\%
confidence interval presented as shadowed regions.
For clarity, only the average value curves are shown after 125 episodes.
}
\label{fig:convergence}
\end{figure}

\begin{figure}[!t]
\centering
\includegraphics[width=0.9\columnwidth]{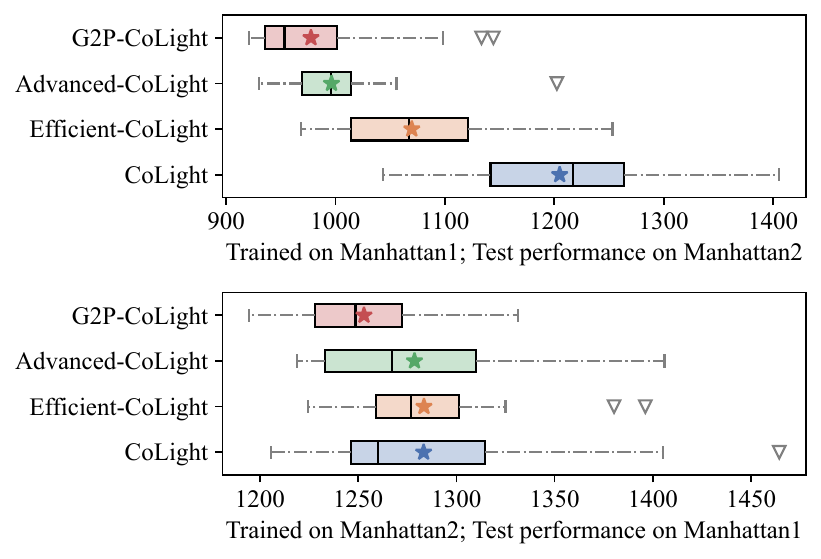}
\caption{The average travel time (the smaller the better) of different algorithms on unseen scenarios.}
\label{fig:generalization}
\end{figure}

\subsection{Performance on Unseen Scenarios}
We test the performance of the trained models of CoLight-based algorithms on unseen scenarios.
To this end, we train a model on one Mahattan scenario and test on the other Manhattan scenario.
The results are presented in Figure \ref{fig:generalization}.
It can be seen that G2P-CoLight stills performs better than other compared methods in unseen Manhattan traffic scenarios.
This indicates that under different traffic conditions within the same road network, G2P-CoLight demonstrates a certain level of generalization ability.

\section{Conclusions}
This paper introduces a entirely new pressure-based algorithm for TSC, called G2P control, which outperforms the SOTA rule-based algorithm [\textbf{RQ1}].
The newly proposed G2P considers both relative and absolute traffic conditions and is proved to be able to stabilize the queue lengths in traffic in this paper [\textbf{RQ2}].
To effectively represent the traffic conditions at an intersection, a new generalized traffic state is designed based on G2P to improve the performance of RL-based methods for TSC [\textbf{RQ3}].
Experimental results show that G2P can significantly improve the performance of RL-based methods as well as data efficiency.

In the future, we will analyze additional unstructured traffic data, such as vehicle routes enabled by the development of the Internet of Vehicles, and develop a more precise traffic state representation based on theoretical justification to further optimize TSC methods.




\bibliography{ref.bib}
\bibliographystyle{icml2025}

\newpage
\appendix
\onecolumn

\renewcommand{\theequation}{A.\arabic{equation}} 

\begin{center}
\Large \textbf{Appendix}
\end{center}
\section{Proof of Theorem 1}
\label{sec:proof}

For each turn movement $\mc{T} = (\mc{R}^{\text{in}},\mc{R}^{\text{out}})$ with its corresponding incoming lanes $\mathcal{L}^{\Rrightarrow}$
 (if $\mc{L}^{\Rrightarrow}$ is an internal lane set, i.e., $ \mathcal{R}^{\text{in}} \in \mathcal{R}_{\text{int}}$), the queue length on lanes $\mc{L}^{\Rrightarrow}$ can be update as follows:
\begin{equation}
\begin{split}
 Q(\mc{L}^{\Rrightarrow})(t+1) &= Q(\mc{L}^{\Rrightarrow})(t) - [C_{\mc{L}^{\Rrightarrow}}(\mc{R}^{\text{out}})(t+1)A(\mc{T})(t)\land Q(\mc{L}^{\Rrightarrow})(t)]\\
 & + \sum_{k}[C_{\mc{R}_k}(\mc{L}^{\Rrightarrow})(t+1)A(\mathcal{R}_k,\mathcal{L}^{\Rrightarrow})(t)\land Q(\mathcal{R}_k)(t)],
\end{split}
\label{eq_x1}
\end{equation}
where $z \land y = \min\{z,y\}$, $Q(\mc{L}^{\Rrightarrow})(t)$ is the length of these homogeneous lanes at the beginning of period $t$. 
$C_{a}(b)(t)$ is non-negative bounded iid random variables with mean value equal to the service or saturation flow rate $c_a(b)$, the second term indicates that up to $C_a(b)(t+1)$ vehicles will leave lane set $a$ and they will be routed to lane set $b$. 
When $A(\mc{T})(t) = 1$, the turn movement $\mc{T} = (\mc{R}^{\text{in}},\mc{R}^{\text{out}})$ is actuated. 

If $\mathcal{L}^{\Rrightarrow} \in \mc{R}^{\text{in}}$ is an entry lane set, it means that external vehicles can enter the road network through these lanes, i.e., $\mathcal{R}^{\text{in}} \in \mathcal{R}_{\text{entry}}$, for each turn movement $\mc{T} = (\mc{R}^{\text{in}},\mc{R}^{\text{out}})$ with its corresponding incoming lanes $\mathcal{L}^{\Rrightarrow}$, the queue length on lanes $\mc{L}^{\Rrightarrow}$ can be update as follows:
\begin{equation}
Q(\mathcal{L}^{\Rrightarrow})(t+1) = Q(\mathcal{L}^{\Rrightarrow})(t) - [C_{\mc{L}^{\Rrightarrow}}(\mc{R}^{\text{out}})(t+1)A(\mc{T})(t)\land Q(\mc{L}^{\Rrightarrow})(t)] + d_{\mc{L}^{\Rrightarrow}}(t+1) 
\label{eq_x2}
\end{equation}
where $d_{\mc{L}^{\Rrightarrow}}(t+1)$ is the exogenous demands with expected rate $d_{\mathcal{L}^{\Rrightarrow}} $ vehicles per period, which $d_{\mathcal{L}^{\Rrightarrow}}$ is the exogenous flow of vehicles in lanes $\mathcal{L}^{\Rrightarrow}$ per period.

Let $A(t) = u^*(Q(t))$ be the phase signal control selected by the G2P control policy (9) in paper.
The update method of queue length is given by \eqref{eq_x1} and \eqref{eq_x2}. 
For any state $Q \in \mathcal{Q}$, let $|Q|^2 = \sum_{\mc{T}}[Q(\mc{L}^{\Rrightarrow})]^2$ be the sum of squares of all the lane lengths. 
We will show that there exist $q < \inf$ and $\epsilon > 0$ such that under $u^*$:
\begin{equation}
    E\left\{|Q(t+1)|^2 - |Q(t)|^2|Q(t) \right\} \leqslant q - \epsilon |Q(t)|,  t= 1,2,...
\label{eq_E}
\end{equation}

Suppose \eqref{eq_E} holds. Taking (unconditional) expectations and summing over $t=1,....,T$ gives:
\begin{equation}
    E|Q(T+1)|^2 - E|Q(1)|^2 \leqslant qT - \epsilon \sum_{t=1}^T E|Q(t)|, 
\end{equation}
and so,
\begin{equation}
    \epsilon \frac{1}{T}E|Q(t)| \leqslant q + \frac{1}{T} E|Q(1)|^2 - \frac{1}{T} E|Q(T+1)|^2 \leqslant q + \frac{1}{T}E|Q(1)|^2,
\end{equation}
which indicates
\begin{equation}
 \frac{1}{T}E|Q(t)| \leqslant \frac{q}{\epsilon}
\end{equation}
it immediately implies the stability condition (16) in paper.
Now our goal is to prove that inequality \eqref{eq_E}  holds.
For convenience, in the subsequent sections, this paper uses $\mc{L}^{\Rrightarrow} \in \mc{R}_{\text{int}}\cup \mc{R}_{\text{entry}}$ to denote $\mc{L}^{\Rrightarrow} \in \mc{R}^{\text{in}}$, where $\mc{R}^{\text{in}} \in \mc{R}_{\text{int}}\cup \mc{R}_{\text{entry}}$, and similarly for other cases. 

\subsection{Proof of \eqref{eq_E}}
Let $A(t) = u(Q(t))$ be any feedback G2P control policy. 
Recall that $Q(t+1)$ is given by \eqref{eq_x1} and \eqref{eq_x2}.
Define the array $\delta = \{\delta(\mc{L}^{\Rrightarrow})\}$ by $\delta = Q(t+1) - Q(t)$, so 

For $\mathcal{L}^{\Rrightarrow} \in \mathcal{R}_{\text{int}}$: 
\begin{equation}
\delta (\mc{L}^{\Rrightarrow}) = - [C_{\mc{L}^{\Rrightarrow}}(\mc{R}^{\text{out}})(t+1)A(\mc{T})(t)\land Q(\mc{L}^{\Rrightarrow})(t)]+ \sum_{k}[C_{\mc{R}_k}(\mc{L}^{\Rrightarrow})(t+1)A(\mathcal{R}_k,\mathcal{L}^{\Rrightarrow})(t)\land Q(\mathcal{R}_k)(t)],
 \label{eq_delta1}
\end{equation}
For $\mathcal{L}^{\Rrightarrow} \in \mathcal{R}_{\text{entry}}$: 
\begin{equation}
\delta (\mc{L}^{\Rrightarrow})  = - [C_{\mc{L}^{\Rrightarrow}}(\mc{R}^{\text{out}})(t+1)A(\mc{T})(t)\land Q(\mc{L}^{\Rrightarrow})(t)]+ d_{\mc{L}^{\Rrightarrow}}(t+1), 
\label{eq_delta2}
\end{equation}
For any two array $Y=\{y(\mathcal{L})\}$ and $Z=\{z(\mathcal{L})\}$, let $Y^TZ = \sum_{\mathcal{T}}Y(\mathcal{L})Z(\mathcal{L})$. So $|Q(t+1)|^2 = Q(t+1)^TQ(t+1)$.
Hence:
\begin{equation}
|Q(t+1)|^2 - |Q(t)|^2 = |Q(t)+\delta|^2 - |Q(t)|^2 = 2Q(t)^T\delta + |\delta|^2 = 2 \sigma + \theta
\end{equation}

\subsubsection{Bound on $\sigma$}
\begin{equation}
    \begin{split}
    & \sigma = Q(t)^T \delta = -\sum_{\mathcal{L}^{\Rrightarrow} \in \mathcal{R}_{\text{int}}} Q(\mc{L}^{\Rrightarrow})(t)[C_{\mc{L}^{\Rrightarrow}}(\mc{R}^{\text{out}})(t+1)A(\mc{T})(t)\land Q(\mc{L}^{\Rrightarrow})(t)] \\
    &+\sum_{\mathcal{R}_k}\sum_{\mathcal{L}^{\Rrightarrow} \in \mathcal{R}_{\text{int}}} Q(\mathcal{L}^{\Rrightarrow})(t) \sum_{k}[C_{\mc{R}_k}(\mc{L}^{\Rrightarrow})(t+1)A(\mathcal{R}_k,\mathcal{L}^{\Rrightarrow})(t)\land Q(\mathcal{R}_k)(t)] \\
    &+ \sum_{\mathcal{L}^{\Rrightarrow} \in \mathcal{R}_{\text{entry}}} Q(\mathcal{L}^{\Rrightarrow})(t)\{- [C_{\mc{L}^{\Rrightarrow}}(\mc{R}^{\text{out}})(t+1)A(\mc{T})(t)\land Q(\mc{L}^{\Rrightarrow})(t)]+ d_{\mc{L}^{\Rrightarrow}}(t+1)\} \\ 
    &= \sum_{\mathcal{L}^{\Rrightarrow} \in \mathcal{R}_{\text{int}}\cup \mathcal{R}_{\text{entry}}}[C_{\mc{L}^{\Rrightarrow}}(\mc{R}^{\text{out}})(t+1)A(\mc{T})(t)\land Q(\mc{L}^{\Rrightarrow})(t)]\left\{-Q(\mathcal{L}^{\Rrightarrow})(t) + \sum Q(\mathcal{R}^{\text{out}})(t)\right\} + \sum_{\mathcal{L}^{\Rrightarrow} \in \mathcal{R}^{\text{entry}}} d_{\mc{L}^{\Rrightarrow}}(t+1)Q(\mathcal{L}^{\Rrightarrow})(t) \\
    &= \sum_{\mathcal{L}^{\Rrightarrow} \in \mathcal{R}_{\text{int}}\cup \mathcal{R}_{\text{entry}}}[C_{\mc{L}^{\Rrightarrow}}(\mc{R}^{\text{out}})(t+1)A(\mc{T})(t)\land Q(\mc{L}^{\Rrightarrow})(t)] \left\{-\left( Q\left(\mc{L}^{\Rrightarrow}\right) -Q^-\left( \mc{L}^{\Rrightarrow} \right)- \sum_{ \mc{L} \in \mc{R}^{\text{out}} } Q\left(\mc{L}\right)\right) -Q^-\left( \mc{L}^{\Rrightarrow} \right) \right\}  \\
    &+ \sum_{\mathcal{L}^{\Rrightarrow} \in \mathcal{R}_{\text{entry}}} d_{\mc{L}^{\Rrightarrow}}(t+1)Q(\mathcal{L}^{\Rrightarrow})(t) \\
    \end{split}
\end{equation}
since $C_a(b)(t+1)$ is independent of $Q(t)$
\begin{equation}
    \begin{split}
    &E\left\{[C_{\mathcal{L}^{\Rrightarrow}}(\mathcal{R}^{\text{out}})(t+1)A(\mathcal{T})(t) \land Q(\mathcal{L}^{\Rrightarrow})(t)]Q(\mathcal{R}^{\text{out}})(t)|Q(t)\right\}   \\
    &=E\left\{E\left\{[C_{\mathcal{L}^{\Rrightarrow}}(\mathcal{R}^{\text{out}})(t+1)A(\mathcal{T})(t)\land Q(\mathcal{L}^{\Rrightarrow})(t)]Q(\mathcal{R}^{\text{out}})(t)|C_{\mathcal{L}^{\Rrightarrow}}(\mathcal{R}^{\text{out}})(t+1),Q(t)\right\}|Q(t)\right\} \\
    &=E\left\{[C_{\mathcal{L}^{\Rrightarrow}}(\mathcal{R}^{\text{out}})(t+1)A(\mathcal{T})(t)\land Q(\mathcal{L}^{\Rrightarrow})(t)]|Q(t)\right\}Q(\mathcal{R}^{\text{out}})(t)
    \end{split}
\end{equation}
By (9) in paper: $\mathbb{P}(\mathcal{T})(t) = \mathbb{P}(\mathcal{T})(Q(t)) = Q\left(\mc{L}^{\Rrightarrow}\right)-Q^-\left( \mc{L}^{\Rrightarrow} \right)- \sum_{ \mc{L} \in \mc{R}^{\text{out}} } Q\left(\mc{L}\right)$, so:
\begin{equation}
\begin{split}
    E\left\{\sigma|Q(t)\right\} &= -\sum_{\mathcal{L}^{\Rrightarrow} \in \mathcal{R}_{\text{int}}\cup \mathcal{R}_{\text{entry}}} E\left\{C_{\mathcal{L}^{\Rrightarrow}}(\mathcal{R}^{\text{out}})(t+1)A(\mc{T})(t)\land Q(\mathcal{L}^{\Rrightarrow})(t)|Q(t)\right\}\mathbb{P}(\mathcal{T})(t) \\
    & -\sum_{\mathcal{L}^{\Rrightarrow} \in \mathcal{R}_{\text{int}}\cup \mathcal{R}_{\text{entry}}} E\left\{C_{\mathcal{L}^{\Rrightarrow}}(\mathcal{R}^{\text{out}})(t+1)A(\mc{T})(t)\land Q(\mathcal{L}^{\Rrightarrow})(t)|Q(t)\right\}Q^-(\mc{L}^{\Rrightarrow})(t) \\
     &+ \sum_{\mathcal{L}^{\Rrightarrow} \in \mathcal{R}_{\text{entry}}}d_{\mathcal{L}^{\Rrightarrow}}Q(\mathcal{L}^{\Rrightarrow})(t) 
\end{split}
\label{eq_sigma}
\end{equation}
Next, using (12) and (13) in paper:
\begin{equation}
\begin{split}
&\sum_{\mathcal{L}^{\Rrightarrow} \in \mathcal{R}_{\text{int}}\cup \mathcal{R}_{\text{entry}},\mc{L} \in \mathcal{R}^{\text{out}}}f_{\mathcal{L}^{\Rrightarrow}}p(\mathcal{L} \mid \mathcal{L}^{\Rrightarrow})\mathbb{P}(\mathcal{T}) = \sum_{\mathcal{L}^{\Rrightarrow}\in \mathcal{R}_{\text{int}}\cup \mathcal{R}_{\text{entry}},\mc{L} \in \mathcal{R}^{\text{out}}}f_{\mathcal{L}^{\Rrightarrow}}p(\mathcal{L} \mid \mathcal{L}^{\Rrightarrow})\left[Q(\mathcal{L}^{\Rrightarrow})- Q^-(\mc{L}^{\Rrightarrow}) - \sum_{\mc{L}\in \mc{R}^{\text{out}}} Q(\mathcal{L})\right] \\
&= \sum_{\mathcal{L}^{\Rrightarrow}\in \mathcal{R}_{\text{int}}\cup \mathcal{R}_{\text{entry}},\mc{L} \in \mathcal{R}^{\text{out}}}f_{\mathcal{L}^{\Rrightarrow}}p(\mathcal{L}\mid \mathcal{L}^{\Rrightarrow})Q(\mathcal{L}^{\Rrightarrow}) - \sum_{\mathcal{L}^{\Rrightarrow}\in \mathcal{R}_{\text{int}}\cup \mathcal{R}_{\text{entry}},\mc{L}\in\mathcal{R}^{\text{out}}} f_{\mc{L}^{\Rrightarrow}}p(\mc{L}\mid\mc{L}^{\Rrightarrow})Q^-(\mc{L}^{\Rrightarrow}) \\
&- \sum_{\mathcal{L}^{\Rrightarrow}\in \mathcal{R}_{\text{int}}\cup \mathcal{R}_{\text{entry}}}\left[ \sum_{\mc{L}\in\mathcal{R}^{\text{out}}}f_{\mathcal{L}^{\Rrightarrow}}p(\mathcal{L}\mid\mathcal{L}^{\Rrightarrow})\right]\sum_{\mc{L}\in\mc{R}^{\text{out}}} Q(\mathcal{L}) \\
&=  \sum_{\mathcal{L}^{\Rrightarrow} \in \mathcal{R}_{\text{int}}\cup \mathcal{R}_{\text{entry}}}f_{\mathcal{L}^{\Rrightarrow}}\left[\sum_{\mc{L}\in\mc{R}^{\text{out}}}p(\mc{L}\mid\mc{L}^{\Rrightarrow})\right]Q(\mathcal{L}^{\Rrightarrow}) -\sum_{\mc{L}\in\mathcal{R}^{\text{out}}\in \mathcal{R}_{\text{int}}}f_{\mathcal{L}} Q(\mathcal{L}) - \sum_{\mathcal{L}^{\Rrightarrow} \in \mathcal{R}_{\text{int}}\cup \mathcal{R}_{\text{entry}}} f_{\mc{L}^{\Rrightarrow}}\left[\sum_{\mc{L}\in\mc{R}^{\text{out}}}p(\mc{L}\mid\mc{L}^{\Rrightarrow})\right]Q^-(\mc{L}^{\Rrightarrow}) \\
& = \sum_{\mathcal{L}^{\Rrightarrow}\in \mathcal{R}_{\text{int}}\cup \mathcal{R}_{\text{entry}}}f_{\mathcal{L}^{\Rrightarrow}} Q(\mathcal{L}^{\Rrightarrow}) -\sum_{\mathcal{L}^{\Rrightarrow}\in \mathcal{R}_{\text{int}}}f_{\mathcal{L}^{\Rrightarrow}} Q(\mathcal{L}^{\Rrightarrow})  - \sum_{\mathcal{L}^{\Rrightarrow}\in \mathcal{R}_{\text{int}}\cup \mathcal{R}_{\text{entry}}} f_{\mc{L}^{\Rrightarrow}}Q^-(\mc{L}^{\Rrightarrow}) \\
& = \sum_{\mathcal{L}^{\Rrightarrow} \in \mathcal{R}_{\text{entry}}} f_{\mathcal{L}^{\Rrightarrow}} Q(\mathcal{L}^{\Rrightarrow})- \sum_{\mathcal{L}^{\Rrightarrow}\in \mathcal{R}_{\text{int}}\cup \mathcal{R}_{\text{entry}}} f_{\mc{L}^{\Rrightarrow}}Q^-(\mc{L}^{\Rrightarrow}) \\
& = \sum_{\mathcal{L}^{\Rrightarrow} \in \mathcal{R}_{\text{entry}}} d_{\mathcal{L}}^{\Rrightarrow}Q(\mathcal{L}^{\Rrightarrow}) - \sum_{\mathcal{L}^{\Rrightarrow}\in \mathcal{R}_{\text{int}}\cup \mathcal{R}_{\text{entry}}} f_{\mc{L}^{\Rrightarrow}}Q^-(\mc{L}^{\Rrightarrow})
\end{split}
\end{equation}
Therefore,
\begin{equation}
\begin{split}
& E\left\{\sigma|Q(t)\right\} = -\sum_{\mathcal{L}^{\Rrightarrow}\in \mathcal{R}_{\text{int}}\cup \mathcal{R}_{\text{entry}},\mathcal{R}^{\text{out}}}E\left\{C_{\mathcal{L}^{\Rrightarrow}}(\mathcal{R}^{\text{out}})(t+1)A(\mathcal{T})(t)\land Q(\mathcal{L}^{\Rrightarrow})(t)|Q(t)\right\}\mathbb{P}(\mathcal{T})(t)  \\ 
& - \sum_{\mathcal{L}^{\Rrightarrow} \in \mathcal{R}_{\text{int}}\cup \mathcal{R}_{\text{entry}},\mathcal{R}^{\text{out}}}E\left\{C_{\mathcal{L}^{\Rrightarrow}}(\mathcal{R}^{\text{out}})(t+1)A(\mathcal{T})(t)\land Q(\mathcal{L}^{\Rrightarrow})(t)|Q(t)\right\}Q^-(\mc{L}^{\Rrightarrow})(t) \\
&+ \sum_{\mathcal{L}^{\Rrightarrow} \in \mathcal{R}_{\text{entry}}} \sum_{\mathcal{R}^{\text{out}}} d_{\mc{L}^{\Rrightarrow}}(t+1)Q(\mathcal{L}^{\Rrightarrow})(t) \\
& = -\sum_{\mathcal{L}^{\Rrightarrow} \in \mathcal{R}_{\text{int}}\cup \mathcal{R}_{\text{entry}},\mathcal{R}^{\text{out}}}E\left\{C_{\mathcal{L}^{\Rrightarrow}}(\mathcal{R}^{\text{out}})(t+1)A(\mathcal{T})(t)\land Q(\mathcal{L}^{\Rrightarrow})(t)|Q(t)\right\}\mathbb{P}(\mathcal{T})(t) \\
& - \sum_{\mathcal{L}^{\Rrightarrow} \in \mathcal{R}_{\text{int}}\cup \mathcal{R}_{\text{entry}},\mathcal{R}^{\text{out}}}E\left\{C_{\mathcal{L}^{\Rrightarrow}}(\mathcal{R}^{\text{out}})(t+1)A(\mathcal{T})(t)\land Q(\mathcal{L}^{\Rrightarrow})(t)|Q(t)\right\}Q^-(\mc{L}^{\Rrightarrow})(t)  \\
& + \sum_{\mathcal{L}^{\Rrightarrow} \in \mathcal{R}_{\text{int}}\cup \mathcal{R}_{\text{entry}},\mathcal{R}^{\text{out}}}f_{\mathcal{L}^{\Rrightarrow}}p(\mathcal{L}\mid \mathcal{L}^{\Rrightarrow})\mathbb{P}(\mathcal{T}) + \sum_{\mathcal{L}^{\Rrightarrow} \in \mathcal{R}_{\text{int}}\cup \mathcal{R}_{\text{entry}},\mathcal{R}^{\text{out}}} f_{\mc{L}^{\Rrightarrow}}Q^-(\mc{L}^{\Rrightarrow}) \\
&= \sum_{\mathcal{L}^{\Rrightarrow} \in \mathcal{R}_{\text{int}}\cup \mathcal{R}_{\text{entry}},\mathcal{R}^{\text{out}}}\left(f_{\mathcal{L}^{\Rrightarrow}}-E\left\{C_{\mathcal{L}}^{\Rrightarrow}(\mathcal{R}^{\text{out}})(t+1)A(\mathcal{T})(t)\land Q(\mathcal{L}^{\Rrightarrow})(t)|Q(t)\right\}\right)\mathbb{P}(\mathcal{T})(t) \\
&+ \sum_{\mathcal{L}^{\Rrightarrow} \in \mathcal{R}_{\text{int}}\cup \mathcal{R}_{\text{entry}},\mathcal{R}^{\text{out}}}\left(f_{\mathcal{L}^{\Rrightarrow}}-E\left\{C_{\mathcal{L}^{\Rrightarrow}}(\mathcal{R}^{\text{out}})(t+1)A(\mathcal{T})(t)\land Q(\mathcal{L}^{\Rrightarrow})(t)|Q(t)\right\}\right)Q^-(\mathcal{L}^{\Rrightarrow})(t) \\
&=\xi_1 + \xi_2
\end{split}
\end{equation}
\subsubsection{Bound on $\xi_1$}
We can have: 
\begin{equation}
\begin{split}
&\xi_1 = \sum_{\mathcal{L}^{\Rrightarrow}\in \mathcal{R}_{\text{int}}\cup \mathcal{R}_{\text{entry}},\mathcal{R}^{\text{out}}}\left(f_{\mathcal{L}^{\Rrightarrow}}-E\left\{C_{\mathcal{L}^{\Rrightarrow}}(\mathcal{R}^{\text{out}})(t+1)A(\mathcal{T})(t)\land Q(\mathcal{L}^{\Rrightarrow})(t)|Q(t)\right\}\right)\mathbb{P}(\mathcal{T})(t) \\
&= \sum_{\mathcal{L}^{\Rrightarrow}\in \mathcal{R}_{\text{int}}\cup \mathcal{R}_{\text{entry}},\mathcal{R}^{\text{out}}}\left[f_{\mathcal{L}^{\Rrightarrow}}-c_{\mathcal{L}^{\Rrightarrow}}(\mathcal{R}^{\text{out}})(t)A(\mc{T})\right]\mathbb{P}(\mathcal{T})(t) \\
&+ \sum_{\mathcal{L}^{\Rrightarrow}\in \mathcal{R}_{\text{int}}\cup \mathcal{R}_{\text{entry}},\mathcal{R}^{\text{out}}}[c_{\mathcal{L}^{\Rrightarrow}}(\mathcal{R}^{\text{out}})-E\{C_{\mathcal{L}^{\Rrightarrow}}(\mathcal{R}^{\text{out}})(t+1)A(\mathcal{T})(t)\land Q(\mathcal{L}^{\Rrightarrow})(t)\}]\mathbb{P}(\mathcal{T})(t) \\
& = \xi_1^1 + \xi_1^2
\end{split}
\end{equation}
with 
\begin{equation}
\xi_1^1 = \sum_{\mathcal{L}^{\Rrightarrow}\in \mathcal{R}_{\text{int}}\cup \mathcal{R}_{\text{entry}},\mathcal{R}^{\text{out}}}\left[f_{\mathcal{L}^{\Rrightarrow}}-c_{\mathcal{L}^{\Rrightarrow}}(\mathcal{R}^{\text{out}})(t)A(\mc{T})\right]\mathbb{P}(\mathcal{T})(t) 
\end{equation}
\begin{equation}
\xi_1^2 = \sum_{\mathcal{L}^{\Rrightarrow}\in \mathcal{R}_{\text{int}}\cup \mathcal{R}_{\text{entry}},\mathcal{R}^{\text{out}}}[c_{\mathcal{L}^{\Rrightarrow}}(\mathcal{R}^{\text{out}})-E\{C_{\mathcal{L}^{\Rrightarrow}}(\mathcal{R}^{\text{out}})(t+1)A(\mathcal{T})(t)\land Q(\mathcal{L}^{\Rrightarrow})(t)\}]\mathbb{P}(\mathcal{T})(t)
\end{equation}

\textbf{Lemma 1}. For all $\mc{T} =(\mc{R}^{\text{in}},\mc{R}^{\text{out}})$ with its corresponding incoming lanes $\mathcal{L}^{\Rrightarrow}$ and $t$,
\begin{equation}
 \xi_1^2 \leqslant \sum_{\mathcal{L}^{\Rrightarrow}\in \mathcal{R}_{\text{int}}\cup \mathcal{R}_{\text{entry}},\mathcal{R}^{\text{out}}} c_{\mathcal{L}^{\Rrightarrow}}(\mathcal{R}^{\text{out}})\overline{C}_{\mathcal{L}^{\Rrightarrow}}(\mathcal{R}^{\text{out}})
\end{equation}

\textbf{Proof.} The function $c \mapsto c \land Q$ is concave in $c$.
Hence:
\begin{equation}
\begin{split}
E\{C_{\mathcal{L}^{\Rrightarrow}}(\mathcal{R}^{\text{out}})(t+1)A(\mathcal{T})(t)\land Q(\mathcal{L}^{\Rrightarrow})(t)|Q(t)\} &\leqslant E\{C_{\mathcal{L}^{\Rrightarrow}}(\mathcal{R}^{\text{out}})(t+1)|Q(t)\} \land Q(\mathcal{L}^{\Rrightarrow})(t) \\
&= c_{\mathcal{L}^{\Rrightarrow}}(\mathcal{R}^{\text{out}}) \land Q(\mathcal{L}^{\Rrightarrow})(t) \\
&\leqslant c_{\mathcal{L}^{\Rrightarrow}}(\mathcal{R}^{\text{out}})
\end{split}
\end{equation}
Let $\overline{C}_{\mathcal{L}^{\Rrightarrow}}(\mathcal{R}^{\text{out}})$ be the maximum value of the random service rate $C_{\mathcal{L}^{\Rrightarrow}}(\mathcal{R}^{\text{out}})(t)$, since $\mathbb{P}(\mathcal{T})(t)\leqslant Q(\mathcal{L}^{\Rrightarrow})(t)$, $A(\mathcal{T})(t)$ is a 0-1 function of $Q(t)$, we can get:
\begin{equation}
\begin{split}
\xi_1^2 &\leqslant \sum_{\mathcal{L}^{\Rrightarrow}\in \mathcal{R}_{\text{int}}\cup \mathcal{R}_{\text{entry}},\mathcal{R}^{\text{out}}}[c_{\mathcal{L}^{\Rrightarrow}}(\mathcal{R}^{\text{out}})-E\{C_{\mathcal{L}^{\Rrightarrow}}(\mathcal{R}^{\text{out}})(t+1)A(\mathcal{T})(t)\land Q(\mathcal{L}^{\Rrightarrow})(t)\}]Q(\mathcal{L}^{\Rrightarrow})(t) \\
&= 
\begin{cases}
    [c_{\mathcal{L}^{\Rrightarrow}}(\mathcal{R}^{\text{out}})-E\{C_{\mathcal{L}^{\Rrightarrow}}(\mathcal{R}^{\text{out}})(t+1)A(\mathcal{T})(t) \land Q(\mathcal{L}^{\Rrightarrow})(t)|Q(t)\}]Q(\mathcal{L}^{\Rrightarrow})(t), & \text{if } Q(\mathcal{L}^{\Rrightarrow}) \leqslant \overline{C}_{\mathcal{L}^{\Rrightarrow}}(\mathcal{R}^{\text{out}}) \\
    0, & \text{if } Q(\mathcal{L}^{\Rrightarrow})> \overline{C}_{\mathcal{L}^{\Rrightarrow}}(\mathcal{R}^{\text{out}})
\end{cases}
\end{split}
\end{equation}

Therefore:
\begin{equation}
0 \leqslant \xi_1^2 \leqslant \sum_{\mathcal{L}^{\Rrightarrow}\in \mathcal{R}_{\text{int}}\cup \mathcal{R}_{\text{entry}},\mathcal{R}^{\text{out}}} c_{\mathcal{L}^{\Rrightarrow}}(\mathcal{R}^{\text{out}})\overline{C}_{\mathcal{L}^{\Rrightarrow}}(\mathcal{R}^{\text{out}})
\end{equation}
which proves the lemma.

\textbf{Lemma 2}. If the G2P control policy $u^*$ is used and $d \in D$ (feasible demand set) there exists $\epsilon >0$ such that:
\begin{equation}
\xi_1^1 \leqslant - \epsilon |Q(t)|
\end{equation} 

\textbf{Proof.} Let $A^*(t) = u^*(Q(t))$, By (17) in paper,
\begin{equation}
\sum A^*(\mc{T})(t)c_{\mc{L}^{\Rrightarrow}}(\mc{R}^{\text{out}})\mathbb{P}(\mc(T))(t) = max \sum A(\mc{T})c_{\mc{L}^{\Rrightarrow}}(\mc{R}^{\text{out}})\mathbb{P}(\mc{T})(t) = max \sum \Sigma_A(\mc{T})c_{\mc{L}^{\Rrightarrow}}(\mc{R}^{\text{out}})\mathbb{P}(\mc{T})(t)
\end{equation}
Since $d$ is a feasible demand (i.e., $d \in D$), there exist $\epsilon > 0$ such that $c_{\mathcal{L}^{\Rrightarrow}}(\mathcal{R}^{\text{out}})\Sigma^+_A(\mc{T}) > f_{\mathcal{L}^{\Rrightarrow}} + \epsilon$ for all $\mc{T}= (\mc{R}^{\text{in}},\mc{R}^{\text{out}})$ with its corresponding incoming lanes $\mathcal{L}^{\Rrightarrow}$ when $ 0 \leqslant \Sigma \leqslant \Sigma^+$.
Hence for fixed $t$, there exists $\Sigma$ such that:
\begin{equation}
\Sigma_A(\mc{T})c_{\mathcal{L}^{\Rrightarrow}}(\mathcal{R}^{\text{out}}) = 
    \begin{cases}
    f_{\mathcal{L}^{\Rrightarrow}}+\epsilon, & \text{if } \mathbb{P}(\mc{T})(t) > 0 \\
    0, & \text{if } \mathbb{P}(\mathcal{T})(t) \leqslant 0
\end{cases}
\end{equation}
Then, 
\begin{equation} 
\begin{split}
    \xi_1^1 &=\sum_{\mathcal{L}^{\Rrightarrow}\in \mathcal{R}_{\text{int}}\cup \mathcal{R}_{\text{entry}},\mathcal{R}^{\text{out}}}[f_{\mathcal{L}^{\Rrightarrow}}-c_{\mathcal{L}^{\Rrightarrow}}(\mathcal{R}^{\text{out}})A^*(\mathcal{T})(t)]\mathbb{P}(\mathcal{T})(t) \\
             &\leqslant \sum_{\mathcal{L}^{\Rrightarrow}\in \mathcal{R}_{\text{int}}\cup \mathcal{R}_{\text{entry}},\mathcal{R}^{\text{out}}} [f_{\mathcal{L}^{\Rrightarrow}}-\Sigma_A(\mathcal{T})c_{\mathcal{L}^{\Rrightarrow}}(\mathcal{R}^{\text{out}})]\mathbb{P}(\mathcal{T})(t) \\
             &= -\epsilon \sum_{\mathcal{L}^{\Rrightarrow}\in \mathcal{R}_{\text{int}}\cup \mathcal{R}_{\text{entry}},\mathcal{R}^{\text{out}}} \mathbb{P}^+(\mathcal{T})(t) + \sum_{\mathcal{L}^{\Rrightarrow}\in \mathcal{R}_{\text{int}}\cup \mathcal{R}_{\text{entry}},\mathcal{R}^{\text{out}}}f_{\mathcal{L}^{\Rrightarrow}}p(\mathcal{R}^{\text{out}}\mid\mathcal{L}^{\Rrightarrow})\mathbb{P}^-(\mathcal{T})(t) \\
             &\leqslant - \epsilon |\mathbb{P}(\mathcal{T})(t)|
\end{split}
\label{eq_xi11}
\end{equation} 
Above $\mathbb{P}^+ = \max\{\mathbb{P},0\},\mathbb{P}^- = \max\{-\mathbb{P},0\}$.

Since the array $\mathbb{P}(\mathcal{T}) = Q(\mathcal{L}^{\Rrightarrow}) - Q^-(\mc{L}^{\Rrightarrow}) - \sum_{\mc{L}\in \mc{R}^{\text{out}}} Q(\mathcal{L})$ is a linear function of the array $Q=\{Q(\mathcal{L}^{\Rrightarrow})\}$, there exists $\zeta > 0$ such that:
\begin{equation}
    \sum |\mathbb{P}(\mathcal{T})| \geqslant \zeta|Q(t)|
\end{equation}
which, together with \eqref{eq_xi11} gives
\begin{equation}
    \xi_1^1 \leqslant -\epsilon \zeta|Q(t)|
\end{equation}
which proves the Lemma 2.

\subsubsection{Bound on $\xi_2$} We can have:
\begin{equation}
\begin{split}
&\xi_2 = \sum_{\mathcal{L}^{\Rrightarrow}\in \mathcal{R}_{\text{int}}\cup \mathcal{R}_{\text{entry}},\mathcal{R}^{\text{out}}}\left(f_{\mathcal{L}^{\Rrightarrow}}-E\left\{C_{\mathcal{L}^{\Rrightarrow}}(\mathcal{R}^{\text{out}})(t+1)A(\mathcal{T})(t)\land Q(\mathcal{L}^{\Rrightarrow})(t)|Q(t)\right\}\right)Q^-(\mathcal{L}^{\Rrightarrow})(t) \\
&= \sum_{\mathcal{L}^{\Rrightarrow} \in \mathcal{R}_{\text{int}}\cup \mathcal{R}_{\text{entry}},\mathcal{R}^{\text{out}}}\left[f_{\mathcal{L}^{\Rrightarrow}}-c_{\mathcal{L}^{\Rrightarrow}}(\mathcal{R}^{\text{out}})(t)A(\mc{T})\right]Q^-(\mathcal{L}^{\Rrightarrow})(t) \\
&+ \sum_{\mathcal{L}^{\Rrightarrow}\in \mathcal{R}_{\text{int}}\cup \mathcal{R}_{\text{entry}},\mathcal{R}^{\text{out}}}[c_{\mathcal{L}^{\Rrightarrow}}(\mathcal{R}^{\text{out}})-E\{C_{\mathcal{L}^{\Rrightarrow}}(\mathcal{R}^{\text{out}})(t+1)A(\mathcal{T})(t)\land Q(\mathcal{L}^{\Rrightarrow})(t)\}]Q^-(\mathcal{L}^{\Rrightarrow})(t) \\
& = \xi_2^1 + \xi_2^2
\end{split}
\end{equation}
with 
\begin{equation}
\xi_2^1 = \sum_{\mathcal{L}^{\Rrightarrow} \in \mathcal{R}_{\text{int}}\cup \mathcal{R}_{\text{entry}},\mathcal{R}^{\text{out}}}\left[f_{\mathcal{L}^{\Rrightarrow}}-c_{\mathcal{L}^{\Rrightarrow}}(\mathcal{R}^{\text{out}})(t)A(\mc{T})\right]Q^-(\mathcal{L}^{\Rrightarrow})(t) 
\end{equation}
\begin{equation}
\xi_2^2 = \sum_{\mathcal{L}^{\Rrightarrow}\in \mathcal{R}_{\text{int}}\cup \mathcal{R}_{\text{entry}},\mathcal{R}^{\text{out}}}[c_{\mathcal{L}^{\Rrightarrow}}(\mathcal{R}^{\text{out}})-E\{C_{\mathcal{L}^{\Rrightarrow}}(\mathcal{R}^{\text{out}})(t+1)A(\mathcal{T})(t)\land Q(\mathcal{L}^{\Rrightarrow})(t)\}]Q^-(\mathcal{L}^{\Rrightarrow})(t) 
\end{equation}

\textbf{Lemma 3}. For all $\mc{T} =  (\mc{R}^{\text{in}},\mc{R}^{\text{out}})$ with its corresponding incoming lanes $\mathcal{L}^{\Rrightarrow}$ and $t$,
\begin{equation}
 \xi_2^2 \leqslant \sum_{\mathcal{L}^{\Rrightarrow} \in \mathcal{R}_{\text{int}}\cup \mathcal{R}_{\text{entry}},\mathcal{R}^{\text{out}}} c_{\mathcal{L}^{\Rrightarrow}}(\mathcal{R}^{\text{out}})\overline{C}_{\mathcal{L}^{\Rrightarrow}}(\mathcal{R}^{\text{out}})
\end{equation}

\textbf{Proof.}
Since $Q^-(\mathcal{L}^{\Rrightarrow})(t)\leqslant Q(\mathcal{L}^{\Rrightarrow})(t)$, $A(\mathcal{T})(t)$ is a 0-1 function of $Q(t)$, we can get:
\begin{equation}
\begin{split}
\xi_2^2 &\leqslant \sum_{\mathcal{L}^{\Rrightarrow}\in \mathcal{R}_{\text{int}}\cup \mathcal{R}_{\text{entry}},\mathcal{R}^{\text{out}}}[c_{\mathcal{L}^{\Rrightarrow}}(\mathcal{R}^{\text{out}})-E\{C_{\mathcal{L}^{\Rrightarrow}}(\mathcal{R}^{\text{out}})(t+1)A(\mathcal{T})(t)\land Q(\mathcal{L}^{\Rrightarrow})(t)\}]Q(\mathcal{L}^{\Rrightarrow})(t) \\
&= 
\begin{cases}
    [c_{\mathcal{L}^{\Rrightarrow}}(\mathcal{R}^{\text{out}})-E\{C_{\mathcal{L}^{\Rrightarrow}}(\mathcal{R}^{\text{out}})(t+1)A(\mathcal{T})(t) \land Q(\mathcal{L}^{\Rrightarrow})(t)|Q(t)\}]Q(\mathcal{L}^{\Rrightarrow})(t), & \text{if } Q(\mathcal{L}^{\Rrightarrow}) \leqslant \overline{C}_{\mathcal{L}^{\Rrightarrow}}(\mathcal{R}^{\text{out}}) \\
    0, & \text{if } Q(\mathcal{L}^{\Rrightarrow})> \overline{C}_{\mathcal{L}^{\Rrightarrow}}(\mathcal{R}^{\text{out}})
\end{cases}
\end{split}
\end{equation}

Therefore:
\begin{equation}
0 \leqslant \xi_2^2 \leqslant \sum_{\mathcal{L}^{\Rrightarrow}\in \mathcal{R}_{\text{int}}\cup \mathcal{R}_{\text{entry}},\mathcal{R}^{\text{out}}} c_{\mathcal{L}^{\Rrightarrow}}(\mathcal{R}^{\text{out}})\overline{C}_{\mathcal{L}^{\Rrightarrow}}(\mathcal{R}^{\text{out}})
\end{equation}
which proves the lemma 3.

\textbf{Lemma 4}. If the G2P control policy $u^*$ is used and $d \in D$ (feasible demand set) there exists $\epsilon >0$ such that:
\begin{equation}
\xi_2^1 \leqslant - \epsilon |Q(t)|
\end{equation} 

\textbf{Proof.} According to Lemma 2, we can get:

there exist $\epsilon > 0$ such that $c_{\mathcal{L}^{\Rrightarrow}}(\mathcal{R}^{\text{out}})\Sigma^+_A(\mc{T}) > f_{\mathcal{L}^{\Rrightarrow}} + \epsilon$ for all $\mc{T}= (\mc{R}^{\text{in}},\mc{R}^{\text{out}})$ with its corresponding incoming lanes $\mathcal{L}^{\Rrightarrow}$ when $ 0 \leqslant \Sigma \leqslant \Sigma^+$.

Then, 
\begin{equation} 
\begin{split}
    \xi_2^1 &=\sum_{\mathcal{L}^{\Rrightarrow}\in \mathcal{R}_{\text{int}}\cup \mathcal{R}_{\text{entry}},\mathcal{R}^{\text{out}}}[f_{\mathcal{L}^{\Rrightarrow}}-c_{\mathcal{L}^{\Rrightarrow}}(\mathcal{R}^{\text{out}})A^*(\mathcal{T})(t)]Q^-(\mathcal{L}^{\Rrightarrow})(t) \\
             &\leqslant \sum_{\mathcal{L}^{\Rrightarrow} \in \mathcal{R}_{\text{int}}\cup \mathcal{R}_{\text{entry}},\mathcal{R}^{\text{out}}} [f_{\mathcal{L}^{\Rrightarrow}}-\Sigma(\mathcal{T})c_{\mathcal{L}^{\Rrightarrow}}(\mathcal{R}^{\text{out}})]Q^-(\mathcal{L}^{\Rrightarrow})(t) \\
             &= -\epsilon\ Q^-(\mathcal{L}^{\Rrightarrow})(t) \\
\end{split}
\label{eq_xi21}
\end{equation} 
Above $Q^-(\mc{L}^{\Rrightarrow})$ can be seen as a linear function of the array $Q=\{Q(\mathcal{L}^{\Rrightarrow})\}$, there exists $\phi > 0$ such that:
\begin{equation}
    Q^-(\mc{L}^{\Rrightarrow}) = \phi |Q(t)|
\end{equation}
which, together with \eqref{eq_xi11} gives
\begin{equation}
    \xi_2^1 \leqslant -\epsilon \phi|Q(t)|
\end{equation}
which proves the Lemma 4.

Combining Lemmas 1, 2, 3, and 4 gives the upper bound:
\begin{equation}
E\{\sigma|Q(t)\} \leqslant - \epsilon |Q(t)| + 2 \cdot \sum_{\mathcal{L}^{\Rrightarrow}\in \mathcal{R}_{\text{int}}\cup \mathcal{R}_{\text{entry}},\mathcal{R}^{\text{out}}}c_{\mathcal{L}^{\Rrightarrow}}(\mathcal{R}^{\text{out}})\overline{C}_{\mathcal{L}^{\Rrightarrow}}(\mathcal{R}^{\text{out}})
\end{equation}

\subsection{Bound on $\theta$}

From \eqref{eq_delta1} and \eqref{eq_delta2}, we can have the bounds:
\begin{equation}
\begin{split}
    |\delta(\mathcal{L}^{\Rrightarrow})| \leqslant \max\{\overline{C}_{\mathcal{L}^{\Rrightarrow}}(\mathcal{R}^{\text{out}}),\sum_{\mathcal{R}_k}\overline{C}_{\mathcal{R}_k}(\mathcal{L}^{\Rrightarrow})\}, \ \ \mathcal{L}^{\Rrightarrow} \in \mathcal{R}_{\text{int}}  \\
    |\delta(\mathcal{L}^{\Rrightarrow})| \leqslant \max\{\overline{C}_{\mathcal{L}^{\Rrightarrow}}(\mathcal{R}^{\text{out}}),\overline{d}_{\mathcal{L}^{\Rrightarrow}}(\mathcal{R}^{\text{out}})\}, \ \ \mathcal{L}^{\Rrightarrow} \in \mathcal{R}_{\text{entry}}
\end{split}
\end{equation}
in which $\overline{d}_{\mathcal{L}^{\Rrightarrow}}(\mathcal{R}^{\text{out}})$ is the maximum value of the random demand $d_{\mathcal{L}^{\Rrightarrow}}(\mathcal{R}^{\text{out}})(t+1)$. Let $\Omega$ be the maximum value of all these bounds. 
Then:
\begin{equation}
    \theta = |\delta|^2 \leqslant B\Omega^2
\end{equation}
in which $B$ is the total number of lanes in the traffic network.

From the above, it can be concluded that: 
\begin{equation}
\begin{split}
    E\{Q|t+1|^2-|Q(t)|^2|Q(t)\} &= E\{2\sigma + \theta |Q(t)\} \leqslant -2 \epsilon |Q(t)|+2 \sum_{\mathcal{L}^{\Rrightarrow} \in \mathcal{R}_{\text{int}}\cup \mathcal{R}_{\text{entry}},\mathcal{R}^{\text{out}}}c_{\mathcal{L}^{\Rrightarrow}}(\mathcal{R}^{\text{out}})\overline{C}_{\mathcal{L}^{\Rrightarrow}}(\mathcal{R}^{\text{out}}) + B\Omega^2 \\
\end{split}
\end{equation}
This proves \eqref{eq_E}, which indicates that G2P control policy $u^*$ is stabilizing on any-homogeneous-lane road networks.

\section{Convergence Curves}
\label{sec:appdx_convergence}
\begin{figure*}[hb]
    \centering

    \subfloat[$\text{Jinan}_1$]{ \includegraphics[width=0.32\textwidth]{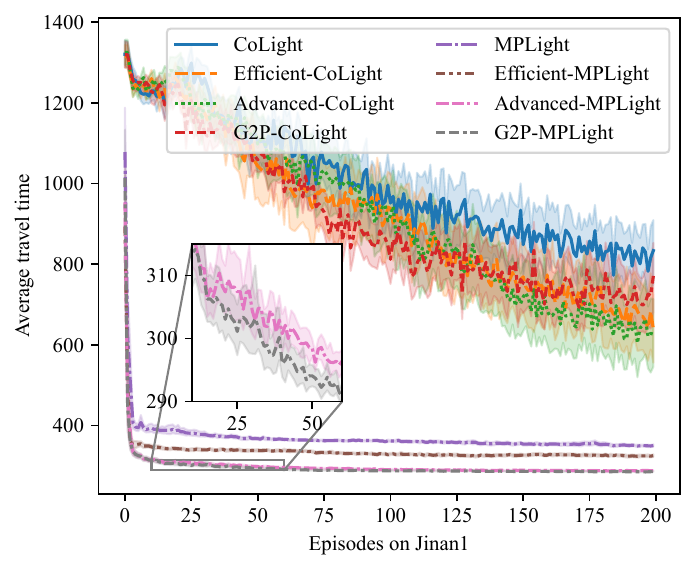} }
    \subfloat[$\text{Jinan}_2$]{ \includegraphics[width=0.32\textwidth]{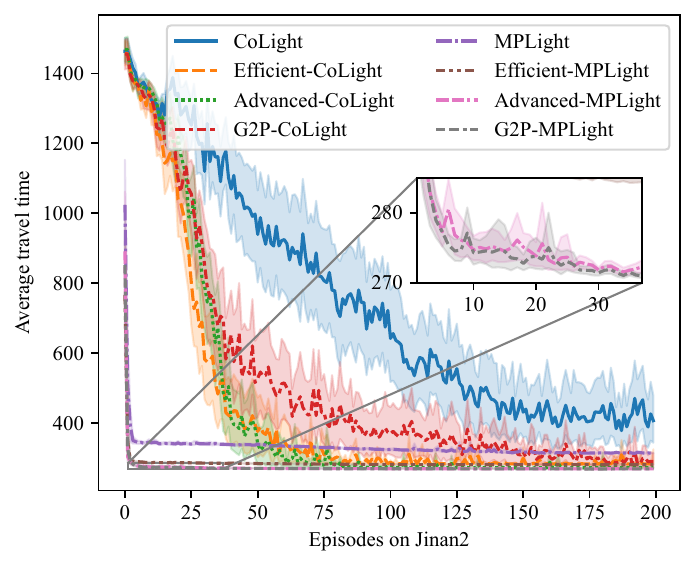} }
    \subfloat[$\text{Jinan}_3$]{ \includegraphics[width=0.32\textwidth]{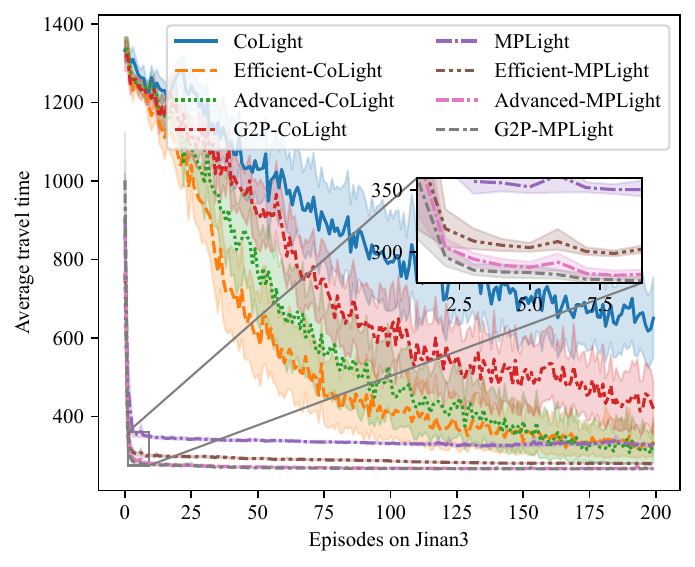} }\\
    \subfloat[$\text{Hangzhou}_1$]{ \includegraphics[width=0.32\textwidth]{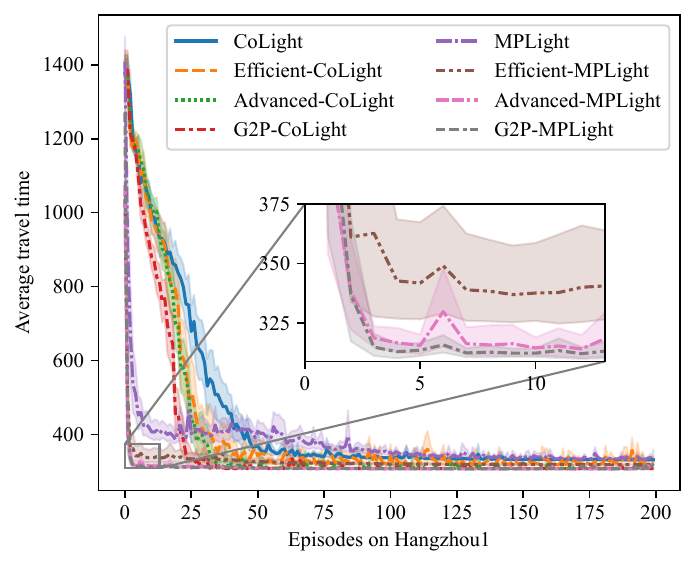} }
    \subfloat[$\text{Hangzhou}_2$]{ \includegraphics[width=0.32\textwidth]{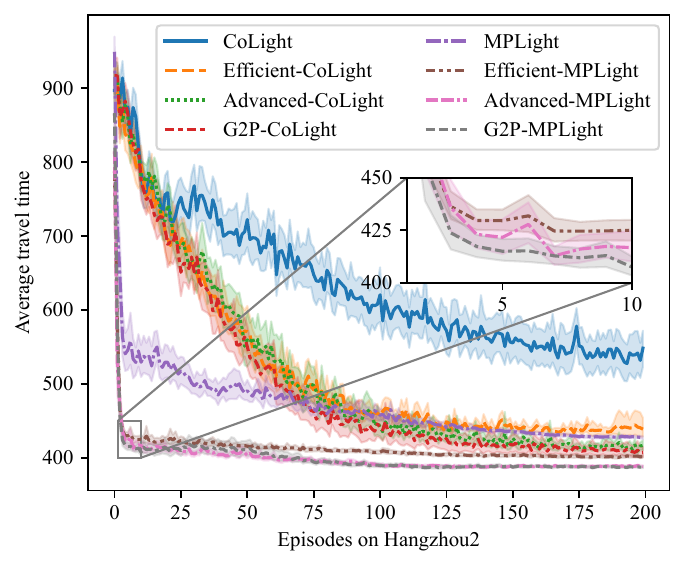} }

    \caption{Convergence curves of different algorithms in different scenarios. For each algorithm we used 30 seeds and obtained the mean-curve with 95\%
confidence interval presented as shadowed regions. MPLight-based methods consistently achieve faster convergence efficiency compared to CoLight-based methods.
As the scale increases (from Jinan to Hangzhou), the advantages of G2P-CoLight become increasingly evident.
}
    \label{fig:terminal_analysis}
\end{figure*}

\end{document}